\documentclass[twoside,11pt]{article}

\usepackage{blindtext}

\usepackage{amsthm}
\usepackage{amsmath}

\usepackage[preprint]{jmlr2e}

\usepackage{multirow}
\usepackage{caption}
\usepackage{subcaption}

\newcommand{\vecT}{\mathrm{vec}}

\newcommand{\dgT}{\mathrm{dg}}
\newcommand{\E}{\mathrm{E}}
\newcommand{\Var}{\mathrm{Var}}
\newcommand{\Cov}{\mathrm{Cov}}
\newcommand{\tr}{\mathrm{tr}}

\theoremstyle{definition}
\newtheorem{dfn}{Definition}

\theoremstyle{plain}
\newtheorem{thm}{Theorem}

\theoremstyle{plain}
\newtheorem{cor}{Corollary}

\theoremstyle{plain}
\newtheorem{prop}{Proposition}

\theoremstyle{plain}
\newtheorem{lem}{Lemma}

\theoremstyle{definition}

\theoremstyle{definition}

\usepackage{lastpage}
\jmlrheading{**}{2023}{1-\pageref{LastPage}}{7/23; Revised */**}{*/**}{**-****}{Luke Duttweiler, Sally W. Thurston, Anthony Almudevar}

\ShortHeadings{Testing Sparsity Assumptions in Bayesian Networks}{Duttweiler, Thurston, and Almudevar}
\firstpageno{1}

\begin{document}

\title{Testing Sparsity Assumptions in Bayesian Networks}

\author{\name Luke Duttweiler \email luke\_duttweiler@urmc.rochester.edu \\
       \addr Department of Biostatistics and Computational Biology\\
       University of Rochester\\
       Rochester, NY 14627, USA
       \AND
       \name Sally W. Thurston \email sally\_thurston@urmc.rochester.edu \\
       \addr Department of Biostatistics and Computational Biology\\
       University of Rochester\\
       Rochester, NY 14627, USA
       \AND
       \name Anthony Almudevar \email anthony\_almudevar@urmc.rochester.edu \\
       \addr Department of Biostatistics and Computational Biology\\
       University of Rochester\\
       Rochester, NY 14627, USA}

\editor{***}

\maketitle

\begin{abstract}
Bayesian network (BN) structure discovery algorithms typically either make assumptions about the sparsity of the true underlying network, or are limited by computational constraints to networks with a small number of variables. While these sparsity assumptions can take various forms, frequently the assumptions focus on an upper bound for the maximum in-degree of the underlying graph $\nabla_G$. Theorem 2 in \cite{duttweiler2023spectral} demonstrates that the largest eigenvalue of the normalized inverse covariance matrix ($\Omega$) of a linear BN is a lower bound for $\nabla_G$. Building on this result, this paper provides the asymptotic properties of, and a debiasing procedure for, the sample eigenvalues of $\Omega$, leading to a hypothesis test that may be used to determine if the BN has max in-degree greater than 1. A linear BN structure discovery workflow is suggested in which the investigator uses this hypothesis test to aid in selecting an appropriate structure discovery algorithm. The hypothesis test performance is evaluated through simulations and the workflow is demonstrated on data from a human psoriasis study.
\end{abstract}


\section{Introduction}

A Bayesian Network (BN) is a type of probabilistic graphical model built on a directed acyclic graph (DAG), which may be used to represent conditional independence or causal relationships among variables. A central focus of much of the research involving BNs is the process of structure discovery. 

Bayesian Network structure discovery focuses on taking a data sample of $p$ variables, which are then treated as vertices in a DAG, and learning which edges exist between what vertices. Many algorithms for structure discovery exist, falling under the general categories of constraint-based and score-based algorithms. Some notable algorithms are the Grow-Shrink and PC-stable constraint-based algorithms of \cite{margaritis2003learning} and \cite{colombo2014order} respectively, and the score-based order-MCMC algorithm of \cite{friedman2003being}.

While structure discovery algorithms approach the problem in vastly different ways, they all make complexity-limiting assumptions about the underlying network, are limited to networks with a fairly low variable count, suffer from high false discovery rates, or some combination of all three of these issues. These issues are all related to the fact that the number of possible DAGs on $p$ variables grows super-exponentially in $p$, leading to important questions about the amount of information available for structure discovery in a dataset where the sample size $n$ is not substantially larger than $p$. 

With these issues in mind, in this paper we develop a novel hypothesis test based on the largest eigenvalue of a transformation of the covariance matrix, which may be used to evaluate the assumption that the largest in-degree of a BN is equal to one. This method is novel not only in that this particular test has not been previously known, but also in that, to the best of our knowledge, this is the first method for learning structural information about a BN from data which does not require the direct use of a structure discovery algorithm. 

We envision the proposed test as a part of a BN structure discovery workflow, in which an investigator first uses our hypothesis test to learn about the complexity of the underlying network. With this knowledge they are then able to select a structure discovery algorithm with a better understanding of which algorithms are appropriate for the task at hand, and in general will have more informed high-level knowledge of the type of network with which they are working.

Section 2 provides background and definitions for linear Bayesian Networks and DAGs, structure discovery, and some matrix algebra that may not be familiar to all readers. Section 3 presents the main theoretical results including giving an overview of the necessary estimation and inference needed for the hypothesis test. Section 4 details a simulation study of the method. Section 5 provides an example of the method's use on real data. Section 6 gives a short discussion on this work along with possible future directions of research. 

Appendix A provides technical details necessary for the estimation and inference on the eigenvalues needed in the hypothesis test. This section includes detailed results for minor advances in eigenvalue shrinkage-estimation, bias-correction, and inference. Appendix B discusses simulations focused on the eigenvalue estimation and inference results of Appendix A.

\section{Background}

In this section we provide definitions and background for linear BNs, structure discovery, and matrix algebra.

\subsection{Linear Bayesian Networks}

\begin{dfn}
We define a graph by $G = (V, E, \xi)$, where $V$ is a set of $p$ vertices and $E$ is a set of edges between those vertices, and $\xi: V\times V \rightarrow \mathbb{R}$ defines a weight function on the set of vertices such that 

\[
\xi(v_i, v_j) \neq 0 \iff (v_i, v_j) \in E.
\]

\noindent We assume for the remainder of this paper that $\xi(v_i, v_i) = 0$ for all $i$.

For any graph $G$ we can define an adjacency matrix $A^{p\times p}$, which is indexed by the vertex set $V$ such that 

\[
A_{ij} = \xi(v_i, v_j).
\]

Finally, if $A_{ij} = A_{ji}$ for all $i,j$ then we say that $G$ is \textit{undirected}, whereas if $A_{ij} \neq 0 \Rightarrow A_{ji} = 0$ for all $i,j$ then we say that $G$ is \textit{directed}.
\end{dfn}

\begin{dfn}
    We say that a \textit{path} of length $s$ exists in $G$ from $v_i$ to $v_j$ if there is a set of distinct vertices $(v^*_0, \dots, v^*_s)$ such that $v_i = v^*_0$, $v_j = v^*_s$ and $\xi(v^*_{l-1}, v^*_l) \neq 0$ for all $l = 1, \dots, s.$ A graph $G$ is called \textit{connected} when for any vertex pair $v_i$ and $v_j$ there exists a path from $v_i$ to $v_j$ or from $v_j$ to $v_i$. 

    When a path of any length exists from $v_i$ back to $v_i$ we call that path a \textit{cycle}. An undirected graph with no cycles is called a \textit{tree} if it is connected, and a \textit{forest} if it is not connected. 

    A directed graph with no cycles is called a \textit{directed acyclic graph} (DAG). Note that if $A$ is the adjacency matrix of a DAG, then we must have that $A$ is nilpotent (ie. there exists a positive integer $s$ such that $A^s = 0$).
\end{dfn}

\begin{dfn}
Let $G$ be a DAG with weight function $\xi$. Then for vertex $v_i$ we define the set of \textit{parents} of $v_i$ as 

\[
pa_G(v_i) = \{v\in V| \xi(v, v_i) \neq 0\},
\]

and the set of \textit{children} of $v_i$ as

\[
ch_G(v_i) = \{v\in V|\xi(v_i, v) \neq 0\}.
\]

We call the size of set $pa_G(v_i)$ the \textit{number of parents} of $v_i$ or the \textit{in-degree} of $v_i$. We denote the \textit{maximum in-degree} (or \textit{maximum parents}) of a graph $G$ with $\nabla_G$, defined as 

\[
\nabla_G = \max_i |pa_G(v_i)|.
\]

Then, following \cite{frydenberg1990chain} we define the \textit{Markov boundary} of $v_i$ to be the set of parents of $v_i$, children of $v_i$ and parents of the children of $v_i$ (not including $v_i$). In set notation,

\[
mb_G(v_i) = pa_G(v_i) \cup ch_G(v_i) \cup \bigg\{v \in pa_G(v_k)|v_k \in ch_G(v_i)\bigg\}\setminus\{v_i\}.
\]
\end{dfn}

\begin{dfn}
Let $G$ be a DAG with adjacency matrix $A$. We define the \textit{moral graph} of $G$, denoted $G_M$, as an undirected graph with adjacency matrix $A_M$ for which

\[
(A_M)_{ij} = 1 \iff v_i \in mb_G(v_j).
\]

Note that, as shown in \cite{duttweiler2023spectral}, $\nabla_G = 1$ if and only if $G_M$ is a tree (or a forest).
\end{dfn}

\begin{dfn}
Let $G$ be a DAG on $p$ vertices with weight function $\xi$ and adjacency matrix $A$. Then, for $v \in V$, let $X = (X_v)$ be a $p \times 1$ random vector indexed by $V$. If we can write 

\[
X = A^TX + \epsilon
\]

where $\epsilon$ is a $p \times 1$ vector of independently distributed error terms with $E[\epsilon_i] = 0$ and $Var(\epsilon_i) = \sigma^2_i > 0,$ we say that $X$ is a \textit{linear Bayesian Network} (linear BN) with respect to $G$. Note that this is equivalent to the definition of a linear structural equation model given in \cite{loh2014high}, and that without loss of generality we assume $X$ is centered on 0.

In order to match standard statistical notation we denote $[A]_{ij} = \beta_{ij}.$

When the vertices in $G$ are ordered topologically such that for all $v_i \in V,$ $pa(v_i) \subseteq \{v_1, \dots, v_{i-1}\},$ then

\[
p(X_i|X_1, \dots, X_{i-1}) = p\big(X_i| \{X_j|v_j \in pa_G(v_i)\}\big).
\]

Thus, a linear BN is a particular case of a Bayesian Network as defined in \cite{pearl2009causality}.
\end{dfn}

There are several matrices related to a linear Bayesian Network that we will be using repeatedly throughout this paper. Below we provide their definitions. 

\begin{dfn}
    Let $X$ be a $p\times 1$ dimensional linear Bayesian Network with adjacency matrix $A$ and error terms $\epsilon$. Let $\Cov(\epsilon) = \Sigma_\epsilon$. Then, observe that $\E[X] = 0$, and therefore the \textit{covariance matrix} of $X$ is defined and denoted as the $p \times p$ matrix
    
    \[\Sigma = \E[XX^T] = \E[(I - A^T)^{-1}\epsilon\epsilon^T(I - A)^{-1}] = (I - A^T)^{-1}\Sigma_\epsilon(I - A)^{-1}.\]

We denote the \textit{inverse covariance matrix} of $X$ by $P = \Sigma^{-1}$, and the diagonal matrix with the diagonal entries of $P$ and zeros everywhere else we denote as $P_d.$ Then we denote the \textit{normalized inverse covariance matrix} of $X$ by $\Omega = P_d^{-1/2}PP_d^{-1/2},$ and the eigenvalues of $\Omega$ by $\lambda_1 > \lambda_2 > \dots > \lambda_p.$ 

\noindent Now, let $X_1, \dots, X_n$ be independent samples of $X$. Then, we denote the \textit{sample covariance matrix} by 

\[
\hat\Sigma = \frac{1}{n}\sum_{i = 1}^n(X_i - \bar{X})(X_i - \bar{X})^T,
\]

\noindent the \textit{sample inverse covariance matrix} by $\hat{P} = \hat\Sigma^{-1},$ denote the diagonal of $\hat{P}$ with $\hat{P}_d$ and the \textit{sample normalized inverse covariance matrix} with $\hat\Omega = \hat{P}^{-1/2}_d\hat{P}\hat{P}^{-1/2}_d.$ Then we denote the eigenvalues of $\hat\Omega$ with $\hat\lambda_1 > \dots > \hat\lambda_p. $
\end{dfn}

\subsection{Structure Discovery and Assumptions}

Let $X_1, \dots, X_n$ be identical and independent samples of a linear BN $X$ with DAG $G$, adjacency matrix $A$, and moral graph $G_M$. BN \textit{structure discovery} is the process in which an investigator uses the samples $X_1, \dots, X_n$ to attempt to learn the true underlying graph $G$, or equivalently, the true adjacency matrix $A$.

While there are a significant number of available structure discovery algorithms, bringing a variety of different approaches to this problem, a common theme among many is an assumption that requires the underlying graph $G$ to be \textit{sparse}. In fact, as demonstrated in \cite{chickering1996learning} and \cite{chickering2004large}, all structure discovery algorithms either make sparsity assumptions or are limited by computational constraints to problems with a small number of nodes. As examples, the PC-stable algorithm of \cite{colombo2014order}, the Grow-Shrink algorithm of \cite{margaritis2003learning}, and the order-MCMC algorithm of \cite{friedman2003being} are all popular BN structure discovery methods that make sparsity assumptions, while the exact discovery method of \cite{silander2012simple} makes no sparsity assumptions, but is only computationally feasible on graphs which have less than 33 vertices.

These sparsity assumptions take various forms depending on the algorithm, but frequently depend (at least partially) on the maximum in-degree of the graph $G$, which we denote with $\nabla_G$. It should be clear then, that any information that can be learned about assumptions on $\nabla_G$ is of interest in the structure discovery process. If it can be reasonably justified from data that $\nabla_G \leq k$ for some constant $k$, this can open the door for faster algorithms that operate in higher dimensions. Particularly relevant to this paper, if the assumption $\nabla_G = 1$ (or equivalently that $G_M$ is a tree or forest) can be reasonably justified, then the polynomial time algorithm presented in \cite{chow1968approximating} will provide a very fast and asymptotically consistent estimate of $G$.

The main results in this paper provide a novel hypothesis test that can test the assumption that $\nabla_G = 1$ ($G_M$ is a tree or forest), giving investigators more information when selecting an algorithm for BN structure discovery. However, before we present the hypothesis test, we provide a few more important definitions. 

\subsection{Some Matrix Algebra}

In this section we present definitions for some matrices and matrix operators that may not be familiar to all readers. Readers interested in a thorough discussion of these operators and matrices should see \cite{magnus2019matrix}.

We begin with definitions of the Kronecker product and the vec operator. 

\begin{dfn}
Let $A$ and $B$ be an $m \times n$ and $p \times q$ matrix respectively. Then the \textit{Kronecker product} of $A$ and $B$ is the $mp \times nq$ matrix defined by

\[
    A \otimes B = \Big[a_{ij}B\Big].
\]
\end{dfn}

\begin{dfn}
Let $A$ be an $m \times n$ matrix, and let $A_{.j}$ be the $j$th column of $A$. Then $\vecT(A)$ is a $mn \times 1$ vector defined by:

\[
    \vecT(A) = \begin{bmatrix}
    A_{.1} \\
    A_{.2} \\
    \vdots \\
    A_{.n}
    \end{bmatrix}
\]
\end{dfn}

We now present two matrices that can be very useful when working with Kronecker products and vectorized matrices. For a more complete exploration of the uses of the commutation matrix and the diagonalization matrix again see \cite{magnus2019matrix} or \cite{neudecker1990asymptotic}.

\begin{dfn}\label{Def: CommutMatrix}
The $mn \times mn$ \textit{commutation matrix}, denoted $K_{m,n}$, is defined  as 

\[
K_{m,n} = \begin{bmatrix}
E_{1,1} & \dots & E_{1,n} \\
\vdots & \ddots & \vdots \\
E_{m,1} & \dots & E_{m,n}
\end{bmatrix},
\]

\noindent where $E_{i,j}$ is a $n \times m$ matrix with a 1 in the $ji$th position and 0s everywhere else. 

For the remainder of the paper we will only be using the $p^2\times p^2$ commutation matrix $K_{p,p}$. Therefore, we suppress the subscript notation and write $K_{p,p} = K.$ Then, following from \cite{neudecker1990asymptotic}, if $A$ and $B$ are both $p\times p$ matrices we have

\[
K(A\otimes B) = (B\otimes A)K.
\]
\end{dfn}

\begin{dfn}\label{Def: DiagonalMatrix}
The \textit{diagonalization matrix}, denoted $D_n$, is defined as the $n^2 \times n^2$ matrix

\[
D_n = \sum_{i}^n(E_{i,i} \otimes E_{i,i})
\]

Again, because we will only ever be using $D_p$ we suppress notation and write this as $D_p = D$. For the interested reader, $D$ is referred to as the diagonalization matrix as 

\[
D\vecT(A) = \vecT(\dgT A),
\]

\noindent where $\dgT A$ is the diagonalized version of $A$ with zeros in each entry except for along the diagonal where the original entries of $A$ are left intact.
\end{dfn}

\noindent This lemma, from \cite{magnus2019matrix} p. 441, gives this useful relationship between the Kronecker product and the $\vecT$ operator.

\begin{lem}[\cite{magnus2019matrix}]\label{Lemma: KroneckerVec}
    For any matrices $A, B, C$ for which the product $ABC$ is defined we have 
    
    \[
        \vecT(ABC) = (C^T \otimes A)\vecT(B).  
    \]
\end{lem}

\noindent Finally, we define a particular version of the Frobenius norm, following \cite{ledoit2004well}.

\begin{dfn}
     Let $A$ be a $p\times p$ matrix. Then, $||A||_F^2 = \tr(A^TA)/p$ denotes the \textit{scaled, squared Frobenius norm}.
\end{dfn}

\section{Main Results}

We now present the hypothesis test that is the main result of this paper. The theory behind the test is presented in section 3.1 and an outline for estimation and inference on the test statistic are presented in sections 3.2 and 3.3 respectively. 

This test should be used as an important part of the BN structure discovery workflow, to confirm or reject an investigator's use of any algorithm that assumes $\nabla_G = 1$. This use is demonstrated in Section 5.

\subsection{The Hypothesis Test}

The hypothesis test developed in this section is based on the largest eigenvalue of the normalized inverse covariance matrix $\Omega$. Theorem \ref{Theorem: MaxEigen} provides the justification for the test and was proven in \cite{duttweiler2023spectral}.

\begin{thm}[\cite{duttweiler2023spectral}]\label{Theorem: MaxEigen}
    Let $X$ be a linear Bayesian Network with moral graph $G_M$ and normalized inverse-covariance matrix $\Omega$, and let $\lambda_1$ be the largest eigenvalue of $\Omega$. Then if $G_M$ is a tree or forest,we must have
    
    \[
    \lambda_1 \leq 2.
    \]
\end{thm}

Thus, in order to determine if $G_M$ is a tree (or equivalently $\nabla_G = 1$) the above result immediately suggests a hypothesis test with null and alternative hypotheses, 

\begin{align*}
    &H_0: \lambda_1 \leq 2 &H_A: \lambda_1  > 2.
\end{align*}

Therefore, defining $\hat\lambda^{c*}_1$ as an appropriate estimator of $\lambda_1$, and $\hat\sigma^2$ as its estimated variance, we calculate our test statistic as

\[
    t = \frac{\hat\lambda^{c*}_1 - 2}{\hat\sigma}.
\]

As we will demonstrate below, we can then expect with a large enough sample size that $t \sim t_{n-p}$ and reject $H_0$ if $t > t^*_{1 -\alpha}$, where $t^*_{1-\alpha}$ is the $1-\alpha$th quantile of the $t_{n-p}$ distribution and $\alpha$ is the pre-selected level of the test. The degrees of freedom value of $n-p$ is used here as our corrected estimator of $\lambda_1$ is a function of all $p$ sample eigenvalues (as will be demonstrated below). 

In the following sections we outline the estimation of $\lambda_1$, and then inference on $\lambda_1$, which includes estimating $\sigma^2.$ The results in the following two sections are proven and explored in more detail in Appendix A, which contains the technical details. 

\subsection{Estimating $\lambda_1$}

\begin{figure}[h]
    \centering
    \includegraphics[width = .7\linewidth]{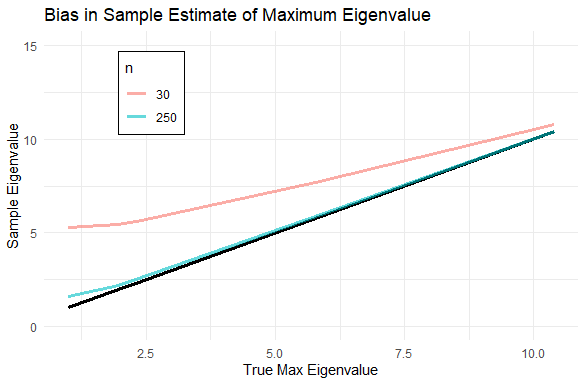}
    \caption{Monte Carlo bias in the sample estimate $\hat\lambda_1$ of the maximum eigenvalue of $\Omega$ with $p = 20$. The x-axis gives the true maximum eigenvalue while the y-axis gives $\hat\lambda_1$. The pink line gives $\hat\lambda_1$ when the sample size is 30, and the blue line gives $\hat\lambda_1$ for a sample size of 250. The black line is a 1:1 line that shows the true value, $\lambda_1$.}
    \label{Fig: sampleBias}
\end{figure}

The primary issue with estimating $\lambda_1$ is the significant bias exhibited by the sample estimate $\hat\lambda_1$ when the true value is small. Figure \ref{Fig: sampleBias} demonstrates this bias by showing a Monte Carlo mean value of $\hat\lambda_1$ as a function of the true value $\lambda_1.$ As can be clearly seen in the figure, when $\lambda_1$ is large there is little sample bias even when the sample size is low, but when $\lambda_1$ is small there is significant positive bias. This is of particular importance in our case as, in order to avoid an inflated Type I error rate, our estimate of $\lambda_1$ must not have bias that increases the expected value to greater than 2 when $\lambda_1 \leq 2.$

Many techniques have been developed to deal with a similar estimation bias presented in the eigenvalues of the covariance matrix. While most of these techniques rely on large-sample properties or Gaussian generative distributions, many share an appealing common property of including a bias correction term that disappears as the sample size grows. For example, with $\lambda_k$ representing the $k$th eigenvalue of $\Sigma$, the estimator 

\[\hat\lambda^A_k = \hat\lambda_k - \frac{1}{n}\sum_{i \neq k}\frac{\hat\lambda_k\hat\lambda_i}{\hat\lambda_k - \hat\lambda_i}\]

\noindent was suggested in \cite{anderson1965asymptotic} as a method of correcting the sample estimate $\hat\lambda_k$, while the estimator 

\[
\hat\lambda^S_k = \frac{\hat\lambda_k}{1 + \frac{1}{n}\sum_{i \neq k}\frac{\hat\lambda_k + \hat\lambda_i}{\hat\lambda_k - \hat\lambda_i}}
\]

\noindent was developed by Stein (as recorded in \cite{muirhead1987developments}) under the assumption that the sample covariance matrix follows a Wishart distribution. More recently, estimators for the eigenvalues of the covariance matrix that function well in a small-sample setting have appeared (for example, see \cite{mestre2008improved}).

Of course, these estimators are all developed for the covariance matrix itself and thus cannot be directly carried over for $\Omega$. However, this reduction of bias when $\lambda_1 \leq 2$ can be achieved by combining a Stein-type shrinkage estimator for $\lambda_1$ with a second-order bias correction based on the estimator proposed in \cite{anderson1965asymptotic}. The specifics for this method are worked out in Appendix A, and simulation studies related to these estimators are provided in Appendix B, but we outline the approach below. Figure \ref{fig: estimatorsBias} is taken from the simulation studies in Appendix B; we will refer to it throughout this section.

We begin with the shrinkage estimation, discuss the second-order bias correction, and then provide the suggested estimator for $\lambda_1$.

\subsubsection{Shrinkage Estimation}

Shrinkage estimators for stabilizing estimates of the covariance matrix in high-dimensional situations have grown greatly in popularity since the introduction of a Stein-type shrinkage estimator based on an objective function using the Frobenius norm in \cite{ledoit2004well}. Further developments of this type of shrinkage estimator continue to be popular due to their computational speed, invariance to permutations, and invertability even when $p > n$ (see \cite{touloumis2015nonparametric}). Of particular interest to us, under particular generative models this family of estimators also does an excellent job of correcting the sample bias of eigenvalues described above.

Significant work has already been done in terms of developing this kind of shrinkage estimation targeted toward the inverse covariance matrix $\Sigma^{-1}$ (see \cite{nguyen2022distributionally} and \cite{bodnar2016direct}), so here we will simply develop a shrinkage estimator for the normalized inverse covariance matrix $\Omega.$ Additionally, we will focus only on the situation where $n > p$ as this aligns with the scope of the rest of the paper. 

Here we present a Stein-type shrinkage estimator for $\Omega$ based on a similar estimator that was proposed for the covariance matrix $\Sigma$ in \cite{ledoit2004well}. As we will show, this estimator immediately provides a shrinkage estimator of $\lambda_1$. 

The development required for the following Theorem is given in Section A.2.

\begin{thm}
    Let $X$ be a $p \times 1$ random vector with finite moments up to the fourth order, and invertible covariance matrix $\Sigma$. Then let $\Omega$ be the normalized inverse covariance matrix of $X$. Now, let $\Omega^*$ be defined by 

    \[
    \Omega^* = (1 - \rho)\hat\Omega + \rho I_p
    \]

    where $\hat\Omega$ is the usual sample estimate of $\Omega$ and $I_p$ is the $p\times p$ identity matrix. Then, the value $\rho^*$ which minimizes the quantity $||\Omega^* - \Omega||_F^2$ is given by 

    \[
    \rho^* = \frac{\tr(\Sigma_{\hat\Omega})}{\tr(\Sigma_{\hat\Omega}) + \tr(\Omega^2) - p},
    \]

    where $\Sigma_{\hat\Omega}$ is the $p^2 \times p^2$ covariance matrix of $\vecT(\hat\Omega),$ and $\Omega^2 = \Omega^T\Omega = \Omega\Omega$.
\end{thm}

Following the language in \cite{ledoit2004well} we note that as $\Omega^*$ depends on the true value and true variance of $\Omega$, it is not a \textit{bona fide} estimator. However, again following \cite{ledoit2004well}, we use plug-in estimators for each component of $\rho^*$, allowing the development of the \textit{bona fide} estimator 

\[
\hat\Omega^* = (1 - \hat\rho^*)\hat\Omega + \hat\rho^* I
\]

\noindent where 

\[
\hat\rho^* = \frac{\widehat{\tr(\Sigma_{\hat\Omega})}}{\widehat{\tr(\Sigma_{\hat\Omega})} + \sum_{i=1}^p\hat\lambda_i^2 - p}.
\]

An estimator of $\Sigma_{\hat\Omega}$ is provided below in Section A.1.4 (Corollary \ref{Corollary: NormInvCovNorm}), immediately giving an estimator of $\tr(\Sigma_{\hat\Omega})$. Additionally, it is important to note that, as shown in Section A.2.2, $\hat\rho^*$ is an asymptotically consistent estimator of $\rho^*$. 

One very useful aspect of this shrinkage estimation is that the eigenvalues of $\hat\Omega^*$ follow the analytical formula 

\[
\hat\lambda_i^* = (1 - \hat\rho^*)\hat\lambda_i + \hat\rho^*,
\]

avoiding the need for additional computation, and ensuring that 
$\hat\lambda^*_1 \geq \hat\lambda^*_2 \geq \dots \geq \hat\lambda^*_p.$ 

\begin{figure}[h]
    \centering
    \includegraphics[width = .9\textwidth]{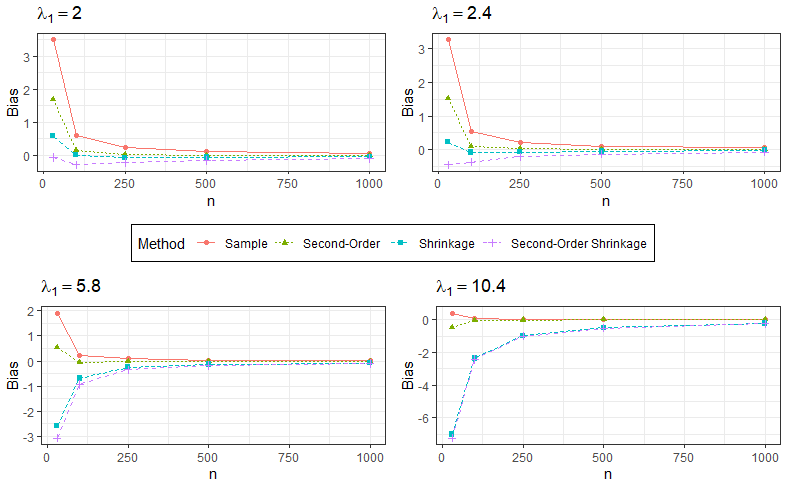}
    \caption{The Monte Carlo bias across simulations estimating $\lambda_1$ where $p = 20$. The simulation is explored in more detail in the Supplementary Materials. Each panel represents a different generative model with the true value of $\lambda_1$ listed above, the x-axis for each gives the sample size, the y-axis gives the bias. The lines/shapes give the different estimation methods where Sample refers to $\hat\lambda_1$, Second-Order to $\hat\lambda_1^c$, Shrinkage to $\hat\lambda_1^*$, and Second-Order Shrinkage to $\hat\lambda_1^{c*}.$}
    \label{fig: estimatorsBias}
\end{figure}

As we can see, $\hat\lambda^*_1$ provides an estimator of $\lambda_1$ which shrinks the sample estimate $\hat\lambda_1$ towards 1. However, as demonstrated in Figure \ref{fig: estimatorsBias}, $\hat\lambda^*_1$ still has a tendency to exhibit positive bias when the true maximum eigenvalue of $\Omega$ is small (close to 2). Thus, we introduce a second-order bias correction which in conjunction with the shrinkage estimation can further reduce the problematic bias, particularly for small $\lambda_1$.

In order to make Figure \ref{fig: estimatorsBias}, we generated data sets of dimension $p = 20$ and sizes $n = 30, 100, 250, 500, 1000$ under 5 different generative models. For model $i$ the data was generated so that $X \sim N(0, \Sigma^{(i)})$, where we chose $\Sigma^{(i)}$ so that the largest eigenvalue of $\Omega^{(i)}$ was $\lambda_1^{(i)} = 2, 2.4, 5.8, 10.4.$ In each situation we generated $B = 300$ data sets and calculated $\hat\lambda, \hat\lambda^c, \hat\lambda^*,$ and $\hat\lambda^{c*}$ from each.

\subsubsection{Second-Order Bias Correction}

Matrix perturbation theory has been applied in several different ways to improve estimates of the eigenvalues of the sample covariance matrix (see \cite{anderson1965asymptotic}, \cite{sakai2000quadratic}, and \cite{fukunaga2013introduction}). This section makes a small adjustment to the application of a well-known matrix perturbation result (see \cite{saleem2015perturbation}) in order to adjust for sample bias in the eigenvalues of $\Omega$. Theorem \ref{Theorem: BiasCorrectionMAIN} presents a result that may be leveraged to further reduce bias in our proposed estimator. The development of this result is given in Section A.3.

\begin{thm}\label{Theorem: BiasCorrectionMAIN}
    Let $X$ be a $p \times 1$ random vector with finite moments up to the fourth order, and invertible covariance matrix $\Sigma$. Then, defining the unit eigenvectors of $\Omega$ as $\omega_1, \dots, \omega_p$ we have
    
    \[
        \E[\hat\lambda_1] = \lambda_1 + \sum_{j = 2}^p\frac{(\omega_j \otimes \omega_1)^T\Sigma_{\hat{\Omega}}(\omega_j \otimes \omega_1)}{\lambda_i - \lambda_1} + \mathcal{O}(n^{-\frac{3}{2}}).
    \]
\end{thm}

Using sample estimates as plug-in values to this equation, this does suggest a bias-corrected estimator of $\lambda_1$. Again noting that an expression for $\hat\Sigma_{\hat\Omega}$ is given in Section A.1.4 (Corollary \ref{Corollary: NormInvCovNorm}), we define this bias-corrected estimator by 

\[
\hat\lambda_1^c = \hat\lambda_1 - \sum_{j \neq 1}\frac{(\hat\omega_j \otimes \hat\omega_1)^T\hat\Sigma_{\hat{\Omega}}(\hat\omega_j \otimes \hat\omega_1)}{\hat\lambda_1 - \hat\lambda_j}.
\]

While $\hat\lambda_1^c$ does have less bias than $\hat\lambda_1$, as demonstrated in Figure \ref{fig: estimatorsBias}, the bias correction does not correctly account for all of the positive bias. Therefore, as suggested above, we will combine shrinkage estimation and the second-order correction to develop our proposed estimator.

\subsubsection{The Proposed Estimator}

Now, observe from Theorem \ref{Theorem: BiasCorrectionMAIN}, given $\hat\rho^*$ we have

\[
\E[\hat\lambda^{*}_1|\hat\rho^*] = (1 - \hat\rho^*)\lambda_1 + (1 - \hat\rho^*)\sum_{j = 2}^p\frac{(\omega_j \otimes \omega_1)^T\Sigma_{\hat{\Omega}}(\omega_j \otimes \omega_1)}{\lambda_1 - \lambda_j} + \hat\rho^* + \mathcal{O}(n^{-\frac{3}{2}}).
\]

\noindent Therefore, for the best estimation of $\lambda_1$ for our suggested hypothesis test, we propose the estimator 

\[
\hat\lambda_1^{c*} = \hat\lambda^*_1 - (1 - \hat\rho^*)\sum_{j = 2}^p\frac{(\hat\omega_j \otimes \hat\omega_1)^T\hat\Sigma_{\hat{\Omega}}(\hat\omega_j \otimes \hat\omega_1)}{\hat\lambda_1 - \hat\lambda_j},
\]

\noindent noting that, as desired

\[
\E[\hat\lambda_i^{c*}|\hat\rho^*] = (1 - \hat\rho^*)\lambda_i + \hat\rho^*  + \mathcal{O}(n^{-\frac{3}{2}}).
\]

As can be seen from Figure \ref{fig: estimatorsBias}, $\hat\lambda_1^{c*}$ is the \textbf{only} estimator which does not exhibit a positive bias for any of the generative models, regardless of sample size. In the current situation, where any positive bias when $\lambda_1  = 2$ will result in an inflated Type I error rate, $\hat\lambda_1^{c*}$ is clearly the best estimator of the set we have examined here. 

Of course, Figure \ref{fig: estimatorsBias} also demonstrates that, at lower sample sizes, $\hat\lambda_1^{c*}$ has a strong \textbf{negative} bias when the true value of $\lambda_1$ is large. This will reduce the power of our hypothesis test at lower sample sizes, but as the bias disappears asymptotically, the power will increase as expected with sample size. 

\subsection{Inference on the Maximum Eigenvalue of $\Omega$}

The asymptotic distributions of the eigenvalues of the sample covariance and correlation matrices have been well studied under various conditions (see\cite{anderson1965asymptotic}, \cite{konishi1979asymptotic}, \cite{van1989elliptical}, \cite{neudecker1990asymptotic}, and \cite{kollo1993asymptotics}). While early development of this theory depended on a multivariate normal generative model, later iterations have discarded this assumption and operate under much more general conditions. Theorem 4 extends these asymptotic distributions to the eigenvalues of $\Omega$ under the multivariate normal case as a specialization of a more general result found in Section A.1. The development and proof of Theorem 4 can be found in Section A.1.

\begin{thm}\label{Theorem: NormInvCovEigenAsymMAIN}
    Let $X \sim N(\mu, \Sigma)$ be a $p \times 1$ random vector with $\Sigma$ non-singular and assume $\hat\Sigma$ is non-singular. Let $\lambda$ be the vector of eigenvalues of $\Omega$ and $\hat\lambda$ be the vector of eigenvalues of $\hat\Omega$.  Then,

\[
    \sqrt{n}(\hat\lambda - \lambda) \xrightarrow[]{D} N(0, R^TU^TVUR)
\]

    \noindent where 

\[
    V = (I_{p^2} + K)(\Sigma \otimes \Sigma),
\]

\[
    U = \Big[\Sigma \otimes \Sigma\Big]^{-1}\Big[(P_d^{-1/2}\otimes P_d^{-1/2}) - \frac12(I_{p^2}+K)(I_p\otimes \Omega P_d^{-1})D\Big],
\]

    \noindent and 

\[
    R = (W\otimes W)J,
\]

\noindent where $W$ is the matrix of orthonormal eigenvectors of $\Omega$ and $J$ is a $p^2\times p$ matrix such that $J = (e_1\otimes e_1 \dots e_p\otimes e_p),$ where $e_i$ is the $p\times 1$ unit vector with a 1 in the $ith$ position and 0s elsewhere.
\end{thm}

The formula given by Theorem \ref{Theorem: NormInvCovEigenAsymMAIN} suggests a plug-in estimator of $\Cov[\hat\lambda] = \Sigma_{\hat\lambda}$ as  $\hat\Sigma_{\hat\lambda} = (n-p)^{-1}\hat{R}^T\hat{U}^T\hat{V}\hat{U}\hat{R}$ where $\hat{V}$ and $\hat{U}$ are defined in Section A.1.4 and $\hat{R} = (\hat{W} \otimes \hat{W})J$ with $\hat{W}$ as the matrix of orthonormal eigenvectors of $\hat\Omega.$ Then, from Theorem \ref{Theorem: NormInvCovEigenAsymMAIN} we see that $\hat\lambda_1$ is asymptotically normal and we can estimate the variance of $\hat\lambda_1$ with $[\hat\Sigma_{\hat\lambda}]_{11}.$

\noindent Now denote the bias correction term with 

\[
\hat{c} = \sum_{j = 2}^p\frac{(\hat\omega_j \otimes \hat\omega_1)^T\hat\Sigma_{\hat{\Omega}}(\hat\omega_j \otimes \hat\omega_1)}{\hat\lambda_1 - \hat\lambda_j},
\]

\noindent and observe that

\begin{align*}
    \Var(\hat\lambda_1^{c*}|\hat\rho^*, \hat{c}) &= \Var\big((1-\hat\rho^*)\hat\lambda_1 + \hat\rho^* + (1 - \hat\rho^*)\hat{c}|\hat\rho^*, \hat{c}\big) \\
    &= (1 - \hat\rho^*)^2\Var(\hat\lambda_1),
\end{align*}

\noindent and since, given $\hat\rho^*$ and $\hat{c}$, $\hat\lambda_1^{c*}$ is simply a rescaling of $\hat\lambda_1$, we know that asymptotically 

\[
\hat\lambda_1^{c*}|\hat\rho^*, \hat{c} \sim N\Big((1 - \hat\rho^*)\lambda_1 + \hat\rho^*, (1 - \hat\rho^*)^2\Var(\hat\lambda_1)\Big).
\]

Of course, a more accurate inference on $\hat\lambda_1^{c*}$ would not be conditional on $\hat\rho^*$ or $\hat{c}$ as these quantities are estimates of constants, not constants themselves. However, the simulation studies below demonstrate that treating these two values as constants in the calculation of variance does not adversely effect the suggested hypothesis test.

With all of these calculations complete we denote $\hat\sigma^2 = (1 - \hat\rho^*)^2\Var(\hat\lambda_1)$ and reiterate from above that

\[
t = \frac{\hat\lambda_1^{c*} - 2}{\hat\sigma} \sim t_{n - p},
\]

\noindent again noting that the degrees of freedom is $n-p$ as $\hat\lambda_1^{c*}$ is a function of all $p$ eigenvalues.

We now provide the results of several simulation studies designed to test our methods performance both when the assumptions are satisfied, and when they are not.

\section{Simulation Studies}

In order to examine the accuracy of our method we ran several simulations under 5 different generative models. The generative models were designed so that model $A$ fulfills all assumptions, models $B$ and $C$ violate the normal errors assumption to differing degrees, and models $D$ and $E$ violate the linearity assumption in different ways. All five models are are described here in detail.

Model $A$ is the most basic model, satisfying all of our assumptions. A BN from model $A$ is a linear Bayesian Network with edge weights sampled from a Gaussian distribution and sampling errors generated from a multivariate normal distribution. Thus for $X$ generated from model $A$ we have 

\[
X = (I - A^T)^{-1}\epsilon
\]

\noindent where $A$ is a nilpotent matrix with non-zero elements generated from a Gaussian distribution and $\epsilon \sim N(0, \Sigma_\epsilon).$ In general we require $\Sigma_\epsilon$ to be a diagonal matrix, but \textbf{do not} require that $\Sigma_\epsilon = \sigma^2 I_p$ for some constant $\sigma^2.$

A BN from model $B$ is still a linear Bayesian Network with edge weights sampled from a Gaussian distribution but violates the assumption that the sampling errors are generated from a multivariate normal distribution and instead assumes that all sampling errors are generated from a $t$ distribution with two degrees of freedom. That is, for each component $X_j$ of $X$ from model $B$ we have

\[
X_j = X_{pa(j)}^T\beta_{pa(j),j} + \epsilon_j
\]

\noindent where $X_{pa(j)}$ is the values of the parents of $X_j$, $\beta_{pa(j),j}$ gives the edge weights of the edges from the parents of $j$ to $j$ (sampled from a Gaussian distribution), and $\epsilon \sim t_2.$

A BN from model $C$ is identical to that of model $B$ but the sampling errors are generated from a $t$ distribution with one degree of freedom so that the errors do not have a finite mean. That is, for each component $X_j$ of $X$ from model $C$ we have

\[
X_j = X_{pa(j)}^T\beta_{pa(j),j} + \epsilon_j
\]

\noindent where $\epsilon_j \sim t_1.$

Model $D$ departs from the linearity assumption. A BN from model $D$ is created by sampling an adjacency matrix with Gaussian edge weights and then assigning each component of the vector $X$ a link function and random distribution from the exponential family. Essentially, data is generated from a BN in model $D$ as if each component is a generalized linear model (GLM) where the linear components of each GLM are determined by the adjacency matrix. Specifically, for each component $X_j$ of $X$ from model $D$ we have 

\[
\E[X_j|X_{pa(j)}] = g^{-1}(X_{pa(j)}^T\beta_{pa(j),j}),
\]

\noindent where the weights $\beta_{pa(j),j}$ are chosen as in models $A - C$. For each BN from model $D$ the conditional distribution was randomly chosen from the set (Gaussian, Bernoulli, Poisson) with probability $(.5, .25, .25)$, and the appropriate canonical link function $g$ was used.

Model $E$ departs from the linearity assumption in a different way. A BN from model $E$ is created by sampling an adjacency matrix with Gaussian edge weights and with non-linear functions from each parent to each child. That is, for each component $X_j$ of $X$ from model $E$, we have 

\[
X_j = \sum_{i \in pa(j)}\beta_{ij}f_i(X_i) + \epsilon_j
\]

\noindent where $f_i$ is a function randomly chosen from the set $(|X_i|, sign(X_i), -2expit(X_i) + 1),$ with equal probability, and $\epsilon_j \sim N(0,1).$

We evaluated the performance of the proposed hypothesis test through several different simulations presented in two subsections below. The first subsection presents the results of hypothesis tests run on data generated from all of the models listed above, with varying values of $\nabla_G$ and varying sample sizes, to demonstrate how the method reacts when the assumptions are met and when the assumptions are violated in different ways. The second subsection presents a detailed power study where data is generated only from model $A$ (so all assumptions are met) and $\nabla_G$ is steadily increased in order to gauge how departure from the null hypothesis will affect statistical power. 

\subsection{Basic Simulations}

Each of the three tables in this subsection reports the results from one simulation, where the simulations differ by each allowing a different value of the maximum number of parents ($\nabla_G$) in the data generating Bayesian Network. In Simulation 1 each BN that is generated has $\nabla_G = 1$, in Simulation 2 each has $\nabla_G = 4$ and in Simulation 3 each BN has $\nabla_G = 8$.

Each simulation was performed by creating 400 BNs, each with $p = 20$ vertices, per generative model, where each BN in a generative model was generated following the process described above. Each BN was then used to generate one dataset of each of the sizes $n = 30,50, 100, 500.$ The proposed hypothesis was performed at a level of $\alpha = .05$ on each data set and the percentage of results which returned a `Reject' result is reported in the tables. 

\begin{table}[ht]
    \centering
    \begin{tabular}{ |c||c||c|c||c|c|| } 
    \hline
    & \multicolumn{5}{|c||}{\textbf{Generative Model}} \\
    \hline
    $n$ & $A$ & $B$ & $C$ & $D$ & $E$ \\ 
    \hline
    \hline
    30 & .05 & .045 & 0 & .045& .04\\ 
    \hline
    50 &  .0525 & .0775 & .055 & .0225 & .0075\\ 
    \hline
    100 & .025 & .04 & \textbf{.1025} & 0& 0\\ 
    \hline
    500 & .03 & .0575 & \textbf{.2125} & 0& 0\\
    \hline
\end{tabular}
    \caption{Percentage of tests with `Reject' result in Simulation 1, where $\nabla_G = 1$ and $H_0$ is true.}
    \label{table: trees}
\end{table}

Table \ref{table: trees} shows the results from Simulation 1, in which each BN has $\nabla_G = 1$, implying that the moral graph of the network is a tree and therefore, that $H_0$ is true. Thus, the results in Table \ref{table: trees} give a Monte Carlo Type I error rate for our hypothesis test under each of the generative models. 

When all assumptions are met (Model $A$) the Type I error rate stays near .05, as desired. Additionally, when the normal errors assumption is relaxed but the errors still have mean 0 (Model $B$), the Type I error rate behaves well. On the other hand, when the errors come from a Cauchy distribution and have no mean (Model $C$), the Type I error rate inflates with the sample size. This error inflation is demonstrated by the bolded values. The two non-linear models ($D$ and $E$) seem to be overly conservative with the Type I error rate going to 0 as the sample size gets larger.

This simulation demonstrates our methods exceptional robustness to non-normal errors. The errors in Model $B$ come from a $t$ distribution with 2 degrees of freedom, which is \textbf{very} far from a normal distribution. Additionally, while the hypothesis test does fail when the errors come from a Cauchy distribution, any preliminary examination of data generated in this way would reveal extreme outliers, demonstrating a need for caution.

\begin{table}[ht]
    \centering
    \begin{tabular}{ |c||c||c|c||c|c|| } 
    \hline
    & \multicolumn{5}{|c||}{\textbf{Generative Model}} \\
    \hline
    $n$ & $A$ & $B$ & $C$ & $D$ & $E$ \\ 
    \hline
    \hline
    30 & .0725 & .055 & .015 & .065& .0375\\ 
    \hline
    50 & .285 & .2975 & .2375 & \textbf{.0225} & \textbf{.025}\\ 
    \hline
    100 & .595 & .68 & .6575 & \textbf{.0075}& \textbf{.0075}\\ 
    \hline
    500 & .995 & .9925 & .9775 & \textbf{.0475}& \textbf{.04}\\
    \hline
\end{tabular}
    \caption{Percentage of tests with `Reject' result in Simulation 2 where $\nabla_G = 4$ and $H_0$ is false.}
    \label{table: nabla4}
\end{table}

Table \ref{table: nabla4} presents the Simulation 2 results. In Simulation 2 $\nabla_G = 4$ for each BN and therefore $H_0$ is false. Thus, the values in Table \ref{table: nabla4} give the Monte Carlo power for the proposed hypothesis test under each model type. All models that satisfy the linearity assumption (Models $A$, $B$ and $C$) have power that increases with sample size, as we would want. However, as shown in bold, the models that do not satisfy linearity ($D$ and $E$) all stay under the level of the test, regardless of sample size.

\begin{table}[ht]
    \centering
    \begin{tabular}{ |c||c||c|c||c|c|| } 
    \hline
    & \multicolumn{5}{|c||}{\textbf{Generative Model}} \\
    \hline
    $n$ & $A$ & $B$ & $C$ & $D$ & $E$ \\ 
    \hline
    \hline
    30 & .0975 & .075 & .025 & .0325& .0525\\ 
    \hline
    50 & .39 & .475 & .505 & .055 & .0775\\ 
    \hline
    100 & .865 & .925 & .885 & .035 & .0475\\ 
    \hline
    500 & 1 & 1 & .98 & \textbf{.2225}& \textbf{.365}\\
    \hline
\end{tabular}
    \caption{Percentage of tests with `Reject' result in Simulation 3 where $\nabla_G = 8$ and $H_0$ is false.}
    \label{table: nabla8}
\end{table}

Table \ref{table: nabla8} presents the Simulation 3 results, that differ from Simulation 2 only in that now $\nabla_G = 8$ instead of 4. Again, all of the models which satisfy linearity behave as expected with increasing power as the sample size grows. Additionally, the models that violate linearity seem to maintain a power value similar to the level of the test, until the sample size gets large enough with $n = 500$. When this happens, as shown in bold, the power of the test does increase although very slowly.

These simulation results seem to suggest that the proposed hypothesis test operates very effectively under the stated assumptions. Additionally, as long as the generating model errors are somewhat constrained (have a finite mean), the hypothesis test works well even when the assumption of normality is violated. However, if the generative model is non-linear, the above results show that the proposed test is severely underpowered, although it does remain conservative. 

\subsection{Power Study Simulations}

In order to evaluate the power of the hypothesis test more thoroughly, we performed a power study. 

To simulate data we began with a single directed graph $G_0$ with $p = 15$ vertices, and for which the moral graph was a tree, implying that $H_0$ was true. $G_0$ was also designed as a complete graph, meaning that every vertex but the founding vertex had one parent. Thus, the addition of any edge into $G_0$ would cause $G_0$ to violate $H_0$.

We then created ten new directed graphs $G^{(1)}_1, \dots, G^{(10)}_1$ by, each time, randomly adding 6 edges into $G_0$. Thus, for each $j = 1,\dots, 10$, $G^{(j)}_1$ was a graph that was 6 edges away from satisfying $H_0.$ Following this, we created ten new graphs $G^{(1)}_2, \dots, G^{(10)}_2$, where for each $j$, $G^{(j)}_2$ was created by randomly adding 6 edges onto $G^{(j)}_1$. We continued this process until we had generated $G^{(1)}_{10}, \dots, G^{(10)}_{10}$, leaving us with 100 directed graphs (excluding $G_0$) which all violate the null hypothesis. 

This data-generation process was used in order to provide some understanding of `distance from $H_0$', by measuring distance using the number of edges away $G_0$ rather than by maximum in-degree in the graph. Of course, as the number of edges in a graph increases, so will $\nabla_G$ for the graph.

\begin{figure}[ht]
    \centering
    \includegraphics[width = .7\textwidth]{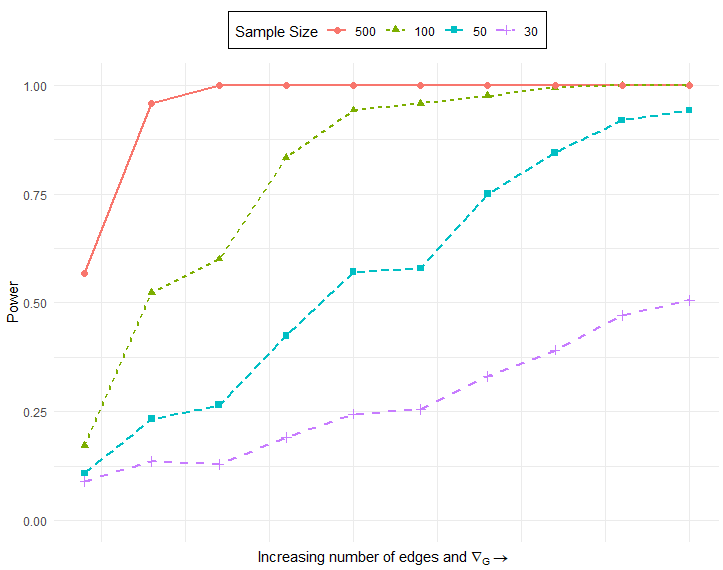}
    \caption{The Monte Carlo power across simulations as a function of sample size $n$ and number of edges. The x-axis gives increasing complexity and maximum in-degree as the graph moves to the right, with each point being an increase in 6 edges from the previous point. The y-axis gives the Monte Carlo estimated power of our proposed hypothesis test.}
    \label{fig: powerstudy}
\end{figure}

Then, for each of the 100 graphs, we generated random edge weights so that each could now function as a linear BN. Then, for each of the four sample sizes $n = 30, 50, 100, 500$, we generated 300 samples from each BN. Our hypothesis test was evaluated on each data set and the results were averaged across graphs with an equal number of edges, giving a Monte Carlo estimate of the power of the test. These results are reported in Figure \ref{fig: powerstudy}.

As would be expected, increasing the sample size improves the power of the test, even when the true graph isn't far from satisfying the null hypothesis. Additionally, as we would hope, the power of the test is increased as the true graph gets farther away from satisfying $H_0$, regardless of sample size. 

\section{Psoriasis Expression Networks}

Psoriasis is a chronic disease in humans characterized by well-defined patches of skin which are scaly and inflamed. There is significant evidence, discussed in \cite{baliwag2015cytokines} and \cite{nickoloff2004recent}, that these `lesions' as they are referred to, are linked to the cytokines expressed in psoriasis skin. As such, there is interest in the gene expression network involving the genes that encode these cytokines.

While a standard approach to an analysis of this network may be to compare gene expressions using correlations or to fit an undirected or BN structure using data, the low sample sizes and high number of parameter estimations required for such a task $\big(p(p-1)/2\big)$ can render these methods ineffective or misleading. Instead, as suggested by the methodology above, we propose here to learn important `global' properties of this network in order to answer important questions with a lesser degree of uncertainty. 

We demonstrate this global property-learning approach on a human psoriasis data set collected as a part of the `Improving Psoriasis Through Health and Well-Being' clinical trial (NIH Project Number: R01AT005082). The data consists of two samples each from $n = 30$ psoriasis patients where one sample contains $p = 22$ gene expression levels from a psoriasis lesion and the other sample contains measurements of the same $p = 22$ genes from a healthy patch of skin. An important and relevant question that may be asked about these two gene expression networks is `Are they the same?' Of course, if the networks are in fact different, it would be best to quantify exactly how they differ, but this is difficult to accomplish with an acceptable degree of certainty at such a low sample size. 

A permutation-based hypothesis test for the equality of two Bayesian Networks is proposed in \cite{almudevar2010hypothesis}. A slight modification of this test (for paired data) allows the user to test the equality of the lesion and non-lesion networks under the assumptions that the networks are Bayesian Networks and have a maximum in-degree of one. While the first of these assumptions must remain an assumption, we are now able to evaluate the validity of the second using the methodology described above. 

We evaluated our hypothesis described above separately on both the lesion and non-lesion data to determine if the assumption that $\nabla_G = 1$ was reasonable for either underlying network. We failed to reject the null on both tests, with test statistic $t = -10.7$ for the non-lesion sample and test statistic $t = -3.1$ for the lesion sample, allowing us to assume $\nabla_G = 1$ for both networks. 

One possible interpretation of this result is that, due to the low sample size, there simply isn't enough information in the data to sustain a more complex estimate with any reasonable degree of certainty. The methods developed in this paper allow us to be more confident in assuming a simple model in order to learn what we can from the data without over-fitting. 

\subsection{Equal Networks?}

We now test the equality of the lesion and non-lesion expression networks. The first network equality test proposed in \cite{almudevar2010hypothesis} is a simple permutation test with $M$ iterations. For each iteration the samples are permuted into two groups of equal size, a tree is fit on each group using the algorithm introduced in \cite{chow1968approximating}, and a score is calculated using this spanning tree. For more details please see \cite{almudevar2010hypothesis}.

Using this hypothesis test (with a slight modification to ensure that the permutations were paired) with $M = 1000$ we calculated a p-value smaller than .001, indicating that the lesion and non-lesion gene expression networks are not identical, which may account for some of the up-regulated cytokines in psoriasis lesions.

\begin{figure}
\centering
\begin{subfigure}{.5\textwidth}
  \centering
  \fbox{\includegraphics[width=.893\linewidth]{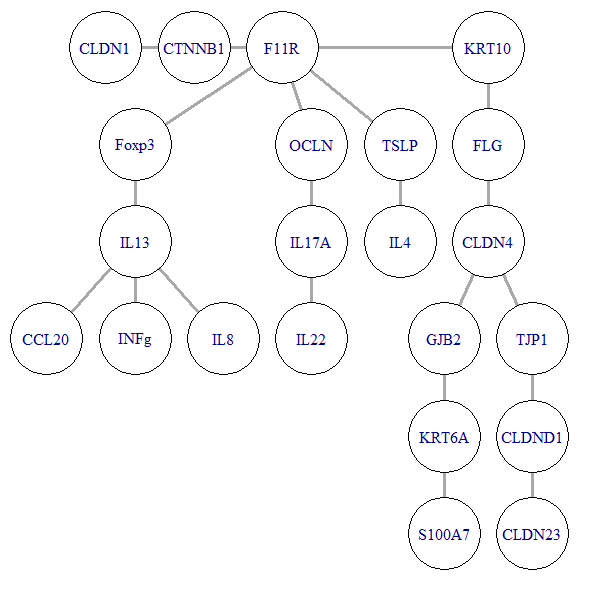}}
  \caption{Non-Lesion Network}
\end{subfigure}%
\begin{subfigure}{.5\textwidth}
  \centering
  \fbox{\includegraphics[width=.9\linewidth]{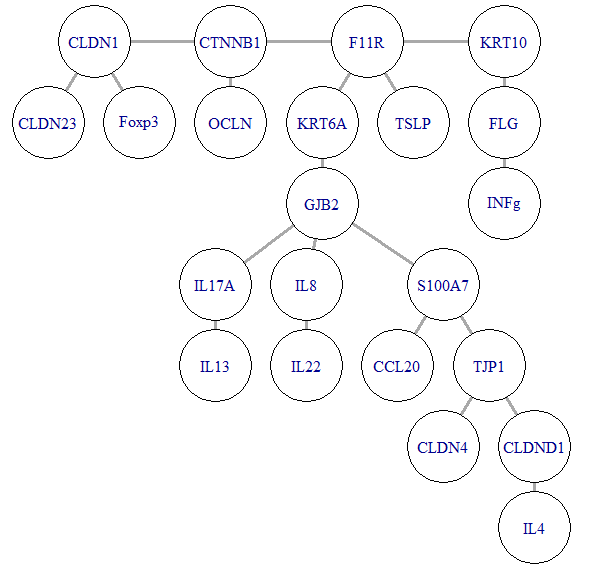}}
  \caption{Lesion Network}
\end{subfigure}
\caption{Networks from the human psoriasis data estimated using the method from \cite{chow1968approximating}. Sub-figure (a) gives the network estimated on the non-lesion portion of the data, while (b) gives the estimation on the lesion portion.}
\label{Figure: ChowLiuFits}
\end{figure}

In order to visualize this difference in networks we fit both networks separately using the minimum spanning tree algorithm from \cite{chow1968approximating}, which again assumes that $\nabla_G = 1$. These networks are presented in Figure \ref{Figure: ChowLiuFits}. 

Additionally we present the networks fit using the TABU algorithm found in \cite{russell2010artificial}. We use the extended BIC of \cite{foygel2010extended} in this score-based algorithm in order to avoid a very dense network. Notice that for this algorithm we relax the assumption that $\nabla_G = 1$.

\begin{figure}
\centering
\begin{subfigure}{.5\textwidth}
  \centering
  \fbox{\includegraphics[width=.9\linewidth]{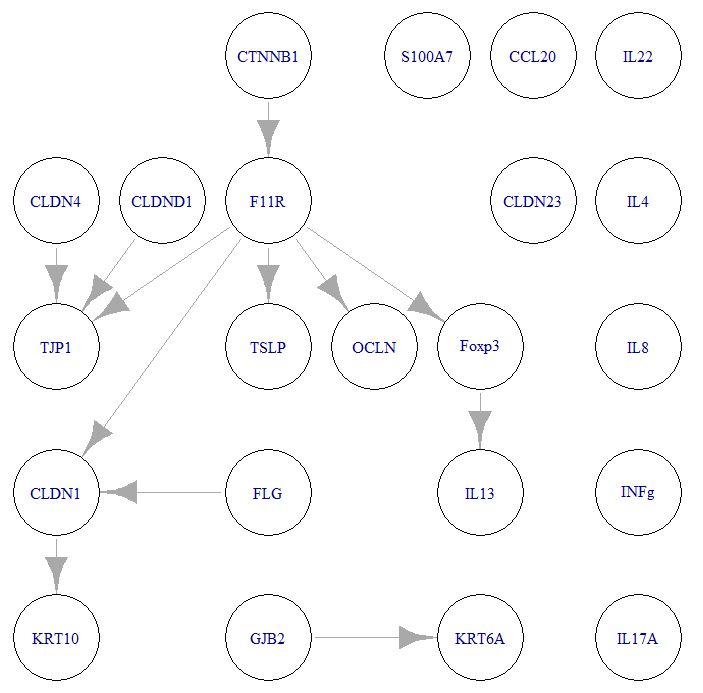}}
  \caption{Non-Lesion Network}
\end{subfigure}%
\begin{subfigure}{.5\textwidth}
  \centering
  \fbox{\includegraphics[width=.906\linewidth]{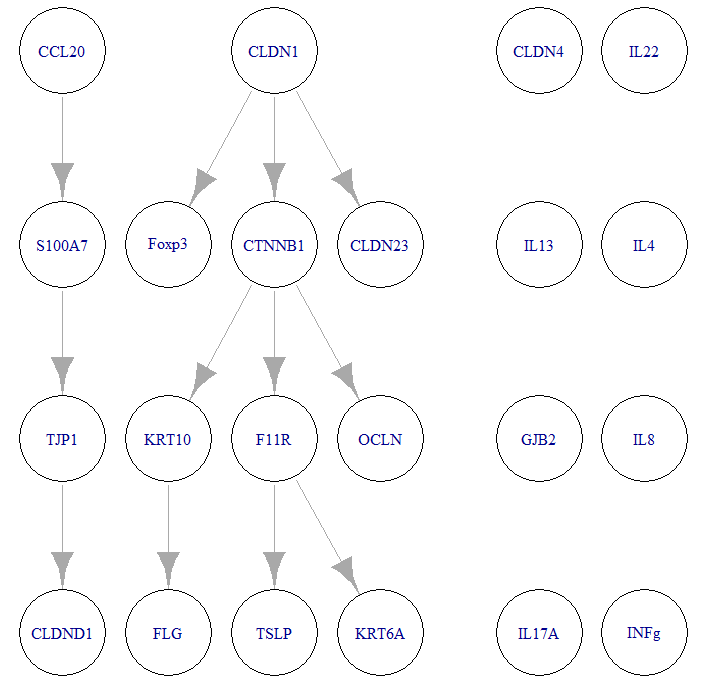}}
  \caption{Lesion Network}
\end{subfigure}
\caption{Networks from the human psoriasis data estimated using the method TABU from \cite{russell2010artificial}. Sub-figure (a) gives the network estimated on the non-lesion portion of the data, while (b) gives the estimation on the lesion portion.}
\label{Figure: TabuFits}
\end{figure}

\section{Discussion and Future Work}

The hypothesis test presented in this paper allows for investigation into a global property of a BN, namely the maximum number of parents, without needing to use an algorithm to fit a model. To our knowledge, this is the first method developed to learn global information about a network without needing to call a particular structure discovery algorithm. This global information can be used to provide justification for the use of certain algorithms and methods, while also helping to give a better overall picture of the underlying network structure.

This is all accomplished without the additional variability that comes from the many different choices made during the estimation of a BN, such as the choice of algorithm or information criterion. Additionally, despite the fact that our method requires an assumption of normality in theory, the simulations demonstrated a strong robustness to the violation of this assumption, except in extreme circumstances. Note however, that this test appears to have two important drawbacks. 

First, because Theorem \ref{Theorem: MaxEigen} does not give an if and only if statement, it is possible to have a BN with $\lambda_1 < 2$ which also has a non-tree moral graph. This can lead to a situation where $H_0$ is true, but the moral graph is actually not a tree ($\nabla > 1$). A second hypothesis test, which has the potential to overcome this problem, was suggested in \cite{duttweiler2023spectral}. This second hypothesis test used all of the eigenvalues instead of just the largest, but the specifics lie beyond the scope of this paper.

Second, as demonstrated in the simulations, the test is severely under-powered in networks with non-linear relationships. While there are many networks of interest that can reasonably take on a linearity assumption (gene expression networks come to mind), there are plentiful examples of networks that cannot. We are hopeful that further research may provide methods to mitigate this problem.

In addition to further research into non-linearity, we hope to work toward estimation and inference on other global properties of Bayesian Networks, such as in-breeding and number of components, without needing to perform an entire structure discovery. Theorem 1 in \cite{duttweiler2023spectral} seems to lay a foundation for further methodology in this vein. 


\acks{The authors would like to acknowledge the incredibly helpful comments and editing of Mr. Luke Rosamilia.

Research reported in this publication was supported by the National Institute of Environmental Health Sciences of the National Institutes of Health (NIH) under award number T32ES007271.  The content is solely the responsibility of the authors and does not necessarily represent the official views of the NIH.

There is no competing interest.}


\newpage

\appendix
\section{}
In the following section we discuss the more technical details required for estimation and inference on the largest eigenvalue of $\Omega$.

\subsection{Asymptotic Distributions}

We begin with a discussion of the asymptotic distributions of $\hat\Sigma^{-1}, \hat\Omega$, and $\hat\lambda$. We first overview necessary matrix derivatives, then apply these derivatives along with the delta method to develop asymptotically normal sample distributions.

\subsubsection{Matrix Function Derivatives}

Here we present three lemmas relating to the derivatives of vectorized matrix functions with respect to the vectorized matrix.

We begin with a derivative of the matrix function which inverts a non-singular matrix. This is a commonly known matrix derivative and thus the proof is omitted.

\begin{lem}\label{Lemma: InverseDeriv}
    Let $A$ be a non-singular, $p \times p$, symmetric matrix. Then, 
    
    \[
        \frac{\partial \vecT(A^{-1})}{\partial \vecT(A)} = -(A^{-1} \otimes A^{-1}).
    \]
\end{lem}

\noindent Next, we show the derivative of the matrix function that normalizes a square matrix. An excellent proof of this Lemma can be found in \cite{neudecker1990asymptotic}.

\begin{lem}[\cite{neudecker1990asymptotic}]\label{Lemma: NormalizedDeriv}
    Let $A$ be a $p\times p$ square, symmetric matrix, $A_d = \dgT A$ and define $\Psi = A_d^{-1/2}AA_d^{-1/2}.$  Then 
    
    \[
    \frac{\partial \vecT(\Psi)}{\partial \vecT(A)} = (A_d^{-1/2} \otimes A_d^{-1/2}) - \frac12(I_{p^2}+K)(I_p\otimes \Psi A_d^{-1})D.
    \]
\end{lem}

\noindent Finally, we present the derivative of the eigenvalues of a symmetric matrix with respect to the vectorized matrix. The full proof can be found in \cite{kollo1993asymptotics}.

\begin{lem}[\cite{kollo1993asymptotics}]\label{Lemma: EigenvalDeriv}
    Let $A$ be a $p \times p$ symmetric matrix with eigenvalues $\lambda_1 > \lambda_2 > \dots > \lambda_p > 0$ and associated orthonormal $p\times 1$ eigenvectors $\omega_1, \dots, \omega_p.$ Let $\Lambda$ be the diagonal matrix with diagonal entries that are the eigenvalues $\lambda_1, \dots, \lambda_p,$ and $W = [\omega_1, \dots, \omega_p].$ Finally, let $\psi\big(\vecT(A)\big)$ be a function of the vector $\vecT(A)$ such that $\psi\big(\vecT(A)\big) = (\lambda_1, \dots, \lambda_p).$ Then,
    
    \[
    \frac{\partial\psi\big(\vecT(A)\big)}{\partial \vecT(A)} = (W\otimes W)J
    \]
    
    \noindent where $J$ is a $p^2\times p$ matrix such that $J = (e_1\otimes e_1 \dots e_p\otimes e_p),$ where $e_i$ is the $p\times 1$ unit vector with a 1 in the $ith$ position and 0s elsewhere.
\end{lem}

\subsubsection{The Covariance Matrix}

Before we can begin deriving the asymptotics for transformations of the covariance matrix, we must have an asymptotic distribution for the covariance matrix itself. The following result, establishing such a distribution under general conditions, was originally published in \cite{neudecker1990asymptotic}. 

\begin{lem}\label{Lemma: CovNorm}
Let $X$ be a $p \times 1$ random vector with finite moments up to the fourth order. Then, 

\[
\sqrt{n}\vecT\big(\hat\Sigma - \Sigma\big) \xrightarrow[]{D} N(0, V)
\]

\noindent with 

\[
V = E[(X-\mu)(X-\mu)^T \otimes (X-\mu)(X-\mu)^T] - (\vecT\Sigma)(\vecT\Sigma)^T.
\]

\end{lem}

In the case that the data is generated from a multivariate normal distribution both \cite{neudecker1990asymptotic} and \cite{van1989elliptical} provide  the following result.

\begin{lem}[\cite{neudecker1990asymptotic} and \cite{van1989elliptical}]\label{Lemma: NormalV}
    Let $X$ be a $p \times 1$ random vector such that $X \sim N(\mu, \Sigma)$. Then, 
    
    \begin{align*}
        V &= E[(X-\mu)(X-\mu)^T \otimes (X-\mu)(X-\mu)^T] - (\vecT\Sigma)(\vecT\Sigma)^T \\
        &= (I_{p^2}+K)(\Sigma \otimes \Sigma).
    \end{align*}
\end{lem}

\subsubsection{The Inverse Covariance Matrix}

Beginning with the inverse covariance matrix, we present the asymptotic distribution of the usual sample estimate under general conditions.

\begin{prop}\label{Proposition: InvCovAsymp}
    Let $X$ be a $p \times 1$ random vector with finite moments up to the fourth order and let $\Sigma$ be non-singular, and assume $\hat\Sigma$ is also non-singular. Recall that we denote $\Sigma^{-1} = P$ and $\hat\Sigma^{-1} = \hat{P}$. Then,

\[
\sqrt{n}\vecT\big(\hat{P} - P\big) \xrightarrow[]{D} N(0, (\Sigma \otimes \Sigma)^{-1}V(\Sigma \otimes \Sigma)^{-1})
\]

\noindent where 

\[
V = E[(X-\mu)(X-\mu)^T \otimes (X-\mu)(X-\mu)^T] - (\vecT\Sigma)(\vecT\Sigma)^T.
\]

\end{prop}

\noindent \textbf{Proof:}

\noindent Let $g$ be a function on vectorized, non-singular, symmetric $p \times p$ matrices defined as 

\[
    g(\vecT(A)) = \vecT(A^{-1}).
\]

\noindent Now, by Lemma \ref{Lemma: InverseDeriv} we have that 

\[
    \frac{\partial g(\vecT(\Sigma))}{\partial\vecT(\Sigma)} = -(\Sigma^{-1} \otimes \Sigma^{-1}) = -(\Sigma \otimes \Sigma)^{-1}.
\]

\noindent Therefore, by the delta method and Lemma \ref{Lemma: CovNorm} we have that 

\[
    \sqrt{n}\Big[g\Big(\vecT\big(\hat\Sigma\big)\Big) - g\Big(\vecT\big(\Sigma\big)\Big)\Big] = \sqrt{n}\vecT\big(\hat\Sigma^{-1} - \Sigma^{-1}\big) \xrightarrow[]{D} N\Big(0, (\Sigma \otimes \Sigma)^{-1}V(\Sigma \otimes \Sigma)^{-1}\Big).\blacksquare
\]

\subsubsection{The Normalized Inverse Covariance Matrix}

Now that we have established the asymptotic distribution of the inverse covariance matrix we do the same with the normalized inverse covariance matrix $\Omega$.

\begin{prop}\label{Proposition: NormInvCovAsymp}
    Let $X$ be a $p \times 1$ random vector with finite moments up to the fourth order, with $\Sigma$ non-singular and assuming $\hat\Sigma$ is non-singular. Then,

\[
\sqrt{n}\vecT\big(\hat\Omega - \Omega\big) \xrightarrow[]{D} N(0, U^TVU)
\]

\noindent where 

\[
V = E[(X-\mu)(X-\mu)^T \otimes (X-\mu)(X-\mu)^T] - (\vecT\Sigma)(\vecT\Sigma)^T,
\]

\noindent and 

\[
U = \Big[\Sigma \otimes \Sigma\Big]^{-1}\Big[(P_d^{-1/2}\otimes P_d^{-1/2}) - \frac12(I_{p^2}+K)(I_p\otimes \Omega P_d^{-1})D\Big],
\]

\noindent where $P_d = \dgT(\Sigma^{-1})$, $K$ is defined in Definition \ref{Def: CommutMatrix} and $D$ is defined in Definition \ref{Def: DiagonalMatrix}.
\end{prop}

\noindent \textbf{Proof:}

\noindent Let $g$ be a function on vectorized, square symmetric matrices such that 

\[
    g(\vecT(A)) = \vecT\big(A_d^{-1/2}AA_d^{-1/2}\big),
\]

\noindent where $A_d = dg(A).$ Then, observe that 

\[
g(\vecT(P)) = \vecT(\Omega) \text{ and } g(\vecT(\hat{P})) = \vecT(\hat\Omega).
\]

\noindent Therefore, by Theorem \ref{Proposition: InvCovAsymp} and the delta method, we must have 

\begin{align*}
    \sqrt{n}\vecT\big(\hat\Omega - \Omega\big) &= \sqrt{n}\big(g(\vecT(\hat{P})) - g(\vecT(P))\big) \\ &\xrightarrow[]{D} N\bigg(0, \frac{\partial g(\vecT(P))}{\partial \vecT(P)}^T(\Sigma \otimes \Sigma)^{-1}V(\Sigma \otimes \Sigma)^{-1}\frac{\partial g(\vecT(P))}{\partial \vecT(P)}\bigg).
\end{align*}

\noindent Then, since by Lemma \ref{Lemma: NormalizedDeriv} we have 

\[
    \frac{\partial g(\vecT(P))}{\partial \vecT(P)} = (P_d^{-1/2}\otimes P_d^{-1/2}) - \frac12(I_{p^2}+K)(I_p\otimes \Omega P_d^{-1})D
\]

\noindent then we must have 

\[
    \frac{\partial g(\vecT(P))}{\partial \vecT(P)}^T(\Sigma \otimes \Sigma)^{-1}V(\Sigma \otimes \Sigma)^{-1}\frac{\partial g(\vecT(P))}{\partial \vecT(P)} = U^TVU
\]

\noindent where 

\[
U = \Big[\Sigma \otimes \Sigma\Big]^{-1}\Big[(P_d^{-1/2}\otimes P_d^{-1/2}) - \frac12(I_{p^2}+K)(I_p\otimes \Omega P_d^{-1})D\Big].
\]

\noindent Therefore, our proof is complete.$\blacksquare$

Although it doesn't simplify anything particularly well, by applying Lemma \ref{Lemma: NormalV} we can obtain the following result in the normal situation.

\begin{cor}\label{Corollary: NormInvCovNorm}
    Let $X \sim N(\mu, \Sigma)$ be a $p\times 1$ random vector and let all other terms be as defined in Proposition \ref{Proposition: NormInvCovAsymp}. Then,
    
    \[
        \sqrt{n}\vecT\big(\hat\Omega - \Omega\big) \xrightarrow[]{D} N(0, U^TV^*U)
    \]
    
    \noindent where 
    
    \[
    V^* = (I_{p^2} + K)(\Sigma \otimes \Sigma).
    \]
\end{cor}

We denote $\Cov[\vecT(\hat\Omega)] = \Sigma_{\hat\Omega}$, and note that an asymptotically unbiased estimator of $\Sigma_{\hat\Omega}$ may be derived by plugging in sample estimators of the various components found in Corollary \ref{Corollary: NormInvCovNorm}. That is, with independent samples $X_1, \dots, X_n$ from $X$, we suggest estimating $\Sigma_{\hat\Omega}$ with

\[
\hat\Sigma_{\hat\Omega} = \frac{1}{n-p}\hat{U}^T\hat{V}\hat{U}
\]

\noindent where

\[
\hat{V} = (I_{p^2} + K)(\hat\Sigma \otimes \hat\Sigma)
\]

\noindent and

\[
\hat{U} = \Big[\hat\Sigma \otimes \hat\Sigma\Big]^{-1}\Big[(\hat{P}_d^{-1/2}\otimes \hat{P}_d^{-1/2}) - \frac12(I_{p^2}+K)(I_p\otimes \hat\Omega \hat{P}_d^{-1})D\Big].
\]

The value $n - p$ in the denominator of the estimator is suggested in order to provide a more conservative estimate of $\Sigma_{\hat\Omega}$ than dividing by $n$, but also to avoid the very restrictive requirement that $n > p(p-1)/2$, which would be required to divide by another justifiable denominator, $n - p(p-1)/2$. For more evidence on the usefulness of this choice, see the simulations in Appendix B. 

\subsubsection{The Eigenvalues of $\Omega$}

\noindent Now we derive the asymptotically normal distribution of the eigenvalues of $\hat\Omega.$

\begin{prop}\label{Proposition: NormInvCovEigenAsym}
    Let $X$ be a $p \times 1$ random vector with finite moments up to the fourth order, with $\Sigma$ non-singular and assuming $\hat\Sigma$ is non-singular. Let $\lambda_1> \dots > \lambda_p > 0$ be the distinct eigenvalues of $\Omega,$ with $\lambda = (\lambda_1, \dots, \lambda_p).$ Then,

\[
    \sqrt{n}(\hat\lambda - \lambda) \xrightarrow[]{D} N(0, R^TU^TVUR)
\]

    \noindent where 

\[
    V = E[(X-\mu)(X-\mu)^T \otimes (X-\mu)(X-\mu)^T] - (\vecT\Sigma)(\vecT\Sigma)^T,
\]

\[
    U = \Big[\Sigma \otimes \Sigma\Big]^{-1}\Big[(P_d^{-1/2}\otimes P_d^{-1/2}) - \frac12(I_{p^2}+K)(I_p\otimes \Omega P_d^{-1})D\Big],
\]

    \noindent and 

\[
    R = (W\otimes W)J,
\]

\noindent where $W$ is the matrix of orthonormal eigenvectors of $\Omega$ and $J$ is a $p^2\times p$ matrix such that $J = (e_1\otimes e_1 \dots e_p\otimes e_p),$ where $e_i$ is the $p\times 1$ unit vector with a 1 in the $ith$ position and 0s elsewhere.
\end{prop}

\noindent\textbf{Proof:}

\noindent Let $g$ be a function on vectorized, square symmetric matrices such that 

\[
    g(\vecT(A)) = \lambda_A,
\]

\noindent where $\lambda_A$ is the vector of eigenvalues of $A$. Then, observe that 

\[
g(\vecT(\Omega)) = \lambda \text{ and } g(\vecT(\hat\Omega)) = \hat\lambda.
\]

\noindent Therefore, by Proposition \ref{Proposition: NormInvCovAsymp} and the delta method, we must have 

\begin{align*}
    \sqrt{n}\vecT\big(\hat\lambda - \lambda\big) &= \sqrt{n}\big(g(\vecT(\hat\Omega)) - g(\vecT(\Omega))\big) \\ &\xrightarrow[]{D} N\bigg(0, \frac{\partial g(\vecT(\Omega))}{\partial \vecT(\Omega)}^TU^TVU\frac{\partial g(\vecT(\Omega))}{\partial \vecT(\Omega)}\bigg).
\end{align*}

\noindent Then, since by Lemma \ref{Lemma: EigenvalDeriv} we have 

\[
    \frac{\partial g(\vecT(\Omega))}{\partial \vecT(\Omega)} = (W\otimes W)J = R.
\]

\noindent then we must have 

\[
    \frac{\partial g(\vecT(\Omega))}{\partial \vecT(\Omega)}^TU^TVU\frac{\partial g(\vecT(\Omega))}{\partial \vecT(\Omega)} = R^TU^TVUR.
\]

\noindent Therefore, our proof is complete.$\blacksquare$

\noindent Finally, using Lemma \ref{Lemma: NormalV} we get the (slight) simplification in the following result. This is presented above in Section 3.3.

\setcounter{thm}{3}
\begin{thm}
    Let $X \sim N(\mu, \Sigma)$ be a $p\times 1$ random vector and let all other terms be as defined in Proposition \ref{Proposition: NormInvCovEigenAsym}. Then
    
    \[
        \sqrt{n}\big(\hat\lambda - \lambda\big) \xrightarrow[]{D} N(0, R^TU^TV^*UR)
    \]
    
    \noindent where 
    
    \[
    V^* = (I_{p^2} + K)(\Sigma \otimes \Sigma).
    \]
\end{thm}
\setcounter{thm}{4}

\subsection{Shrinkage Estimation}

This section focuses on the technical details of the shrinkage estimation approach and is split into three subsections. In the first, we present the minimizer $\Omega^*$ of a Frobenius norm-based objective function of $\Omega$ and $\hat\Omega$. In the second, we develop this function into a true estimator of $\Omega$ using $N$-consistent plug-in estimators of the different terms found in $\Omega^*$. Finally, in the third subsection, we explore the impact this has on the estimated eigenvalues. 

\subsubsection{Minimizing an Objective Function}

Following \cite{ledoit2004well} we propose a Stein-type shrinkage estimator of the normalized inverse covariance matrix $\Omega$ as

\[\Omega^* = (1-\rho)\hat\Omega + \rho I\]

\noindent such that $E[||\Omega^* - \Omega||_F^2]$ is minimized over $\rho$. 

It should be clear that the suggested form of $\Omega^*$ is shrinking the sample matrix $\hat\Omega$ toward a target matrix, namely the identity matrix $I$. Shrinkage estimation for the covariance and inverse-covariance matrices requires the researcher to spend a great deal of thought choosing a target matrix of the correct form. However, when dealing with the normalized inverse-covariance matrix no such choice is necessary as the diagonal of $\Omega$ must be all ones by definition. Hence, we only need focus on shrinking toward the identity. 

Under the assumptions that all expectations involved exist, and that $E[\hat\Omega] = \Omega$ (which is true asymptotically as shown in Proposition \ref{Proposition: NormInvCovAsymp}), this minimization problem has already been solved in \cite{ledoit2004well}. We present a rewording of their minimization solution here, without proof. 

\begin{lem}[\cite{ledoit2004well}]\label{Lemma: L&WolfMinimization}
    Consider the optimization problem 

    \[\min_\rho[||\Omega^* - \Omega||_F^2]\]

    \noindent where $\Omega^* = (1-\rho)\hat\Omega + \rho I$. The solution is given by 

    \[
    \rho^* = \frac{\E\big[||\hat\Omega - \Omega||_F^2\big]}{\E\big[||\hat\Omega - I||_F^2]}.
    \]
\end{lem}

\noindent This Lemma is easily verified through the proof in \cite{ledoit2004well} by substituting $\Omega$ for $\Sigma$ and setting $v = 1.$

Of course, since $\rho^*$ is a function of unobservable quantities we are unable to use this directly to estimate $\Omega^*.$ However, now following the development of similar theory (in a non-parametric setting) given in \cite{touloumis2015nonparametric}, we present $\rho^*$ as a function of several estimable quantities, which may be individually estimated, leading to a plug-in estimator of $\rho^*$. This Theorem was presented above without proof in Section 3.2.

\setcounter{thm}{1}
\begin{thm}\label{Theorem: rhoStarEstimable}
    Let $X$ be a $p \times 1$ random vector and let $X_1, \dots, X_n$ be independent samples of $X$. Define $\Sigma_{\hat\Omega}$ as the variance of $\vecT(\hat\Omega)$. Consider the optimization problem determining

    \[\min_\rho[||\Omega^* - \Omega||_F^2]\]

    \noindent where $\Omega^* = (1-\rho)\hat\Omega + \rho I$. The solution is given by 

    \[
    \rho^* = \frac{\tr(\Sigma_{\hat\Omega})}{\tr(\Sigma_{\hat\Omega}) + \sum_{i=1}^p\lambda_i^2 - p}.
    \]
\end{thm}
\setcounter{thm}{4}

\noindent \textbf{Proof:}

\noindent First, observe from \cite{ledoit2004well} (Lemma 2.1) that we have 

\[
\E\big[||\hat\Omega - \Omega||^2_F\big] = \E\big[||\hat\Omega - I||^2_F\big] - ||\Omega - I||^2_F.
\]

\noindent Therefore from Lemma \ref{Lemma: L&WolfMinimization} we have 

\[
\rho^* = \frac{\E\big[||\hat\Omega - I||^2_F\big] - ||\Omega - I||^2_F}{\E\big[||\hat\Omega - I||^2_F\big]}.
\]

\noindent Then, since 

\[
||\Omega - I||^2_F = \frac{\tr(\Omega^2) - 2\tr(\Omega) + \tr(I)}{p} = \frac{\tr(\Omega^2) - p}{p},
\]

\noindent and 

\[
\E\big[||\hat\Omega - I||^2_F\big] = \frac{\E\big[\tr(\hat\Omega^2) - 2\tr(\hat\Omega) + \tr(I)\big]}{p} = \frac{\E[\tr(\hat\Omega^2)] - p}{p} = \frac{\tr(\Sigma_{\hat\Omega}) + \tr(\Omega^2) - p}{p},
\]

\noindent then we must have that 

\begin{align*}
    \rho^* &= \frac{\tr(\Sigma_{\hat\Omega}) + \tr(\Omega^2) - p - \tr(\Omega^2) + p}{\tr(\Sigma_{\hat\Omega}) + \tr(\Omega^2) - p} \\
    &= \frac{\tr(\Sigma_{\hat\Omega})}{\tr(\Sigma_{\hat\Omega}) + \tr(\Omega^2) - p}.
\end{align*}

\noindent Finally, observe that $\tr(\Omega^2) = \sum_{i=1}^p\lambda_i^2,$ giving our result. $\blacksquare$ 

\subsubsection{A Shrinkage Estimator for $\Omega$}

We now discuss the estimation of $\rho^*$, approaching the estimation through the plug-in estimation of each component of $\rho^*$. As each is a function of the covariance matrix $\Sigma$ we also demonstrate that the sample estimates are consistent in $n$, assuming that the data is generated from a normal distribution. 

\noindent Recall from Proposition \ref{Proposition: NormInvCovAsymp} that we have $\Sigma_{\hat\Omega} = U^TVU/n$, where 

\[
V = E[(x-\mu)(x-\mu)^T \otimes (x-\mu)(x-\mu)^T] - (\vecT\Sigma)(\vecT\Sigma)^T,
\]

\noindent and 

\[
U = \Big[\Sigma \otimes \Sigma\Big]^{-1}\Big[(P_d^{-1/2}\otimes P_d^{-1/2}) - \frac12(I+K)(I\otimes \Omega P_d^{-1})D\Big].
\]

\noindent Additionally, following from Lemma \ref{Lemma: NormalV}, we know that if $X \sim N(\mu, \Sigma)$ then $\Sigma_{\hat\Omega} = U^TV^*U/n$, where 

\[V^* = (I+K)(\Sigma\otimes\Sigma).\]

\noindent Define the plug-in estimator $\widehat{\tr(\Sigma_{\hat\Omega})} = \tr(\hat{U}^T\hat{V}^*\hat{U})/n$ where 

\[\hat{V}^* = (I+K)(\hat\Sigma\otimes\hat\Sigma)\]

\noindent and 

\[
\hat{U} = \Big[\hat\Sigma \otimes \hat\Sigma\Big]^{-1}\Big[(\hat{P}_d^{-1/2}\otimes \hat{P}_d^{-1/2}) - \frac12(I+K)(I\otimes \hat\Omega \hat{P}_d^{-1})D\Big].
\]

\noindent We now have a simple, but important result.

\begin{thm}\label{Theorem: rhoStarConsistent}
    Let $X$ be a $p\times p$ random vector such that $X \sim N(\mu, \Sigma)$, and define all other terms as given in Proposition \ref{Proposition: NormInvCovAsymp}. Next, define $0 < \rho^* < 1$ as in Theorem \ref{Theorem: rhoStarEstimable}, and 

    \[
    \hat\rho^* = \frac{\widehat{\tr(\Sigma_{\hat\Omega})}}{\widehat{\tr(\Sigma_{\hat\Omega})} + \sum_{i=1}^p\hat\lambda_i^2 - p}.
    \]

    \noindent Then, $\hat\rho^*$ is an $n$-consistent estimator of $\rho^*$.
\end{thm}

\noindent \textbf{Proof:}

Recall that, as defined in Proposition \ref{Proposition: NormInvCovAsymp}, $\hat\Sigma$ is an $n$-consistent estimator for $\Sigma$. Then, observe that we can easily define a function $g$ such that $\rho^* = g(\Sigma)$ while $\hat\rho^* = g(\hat\Sigma).$  Finally, observing that $g$ must be continuous we have by the continuous mapping theorem that $\hat\rho^*$ is a consistent estimator for $\rho^*.\blacksquare$

Notice that in our plug-in estimator for the variance of $\hat\Omega$ we divide by $n$. While with increasing $n$ and constant $p$ all three of these options are asymptotically equivalent, there is also an argument to be made for dividing by $n-p$ or $n- p(p-1)/2.$ The simulations in the Supplementary Material provide evidence for dividing by $n-p$ as we are assuming that $n > p$ but not necessarily $n > p(p-1)/2$, and as dividing by $n-p$ typically gives a more `conservative' estimate for $\rho^*$. 

In any case, this estimator $\hat\rho^*$ provides a method for an improved estimation of $\Omega$ which exhibits several nice properties. Of particular interest to us, estimates of the eigenvalues provided through this shrinkage technique are very easy to compute, exhibit less variance, and demonstrate a significant decrease in small-sample bias in some cases. The next subsection explores the theoretical implications of this shrinkage on the eigenvalues.

\subsubsection{Shrinking the Eigenvalues}

Observe that once we have a shrinkage estimate of $\Omega$, call this estimate $\hat\Omega^*$, we are immediately able to calculate the shrinkage eigenvalues, without computing an eigen-decomposition of $\hat\Omega^*$. 

\noindent Let $\hat\lambda_i$ be the $i$th eigenvalue of $\hat\Omega$. Then for the corresponding eigenvector $\hat\omega_i$ we have 

\begin{align*}
    \hat\Omega\hat\omega_i = \hat\lambda_i\hat\omega_i &\Rightarrow (1-\hat\rho^*)\hat\Omega\hat\omega_i = (1 - \hat\rho^*)\hat\lambda_i\hat\omega_i \\
    &\Rightarrow (1-\hat\rho^*)\hat\Omega\hat\omega_i + \hat\rho^*I\hat\omega_i = (1 - \hat\rho^*)\hat\lambda_i\hat\omega_i + \hat\rho^*\hat\omega_i \\
    &\Rightarrow \big[(1 - \hat\rho^*)\hat\Omega + \hat\rho^*I\big]\hat\omega_i = \big[(1 - \hat\rho^*)\hat\lambda_i + \hat\rho^*\big]\hat\omega_i \\
    &\Rightarrow \hat\Omega^*\hat\omega_i = \hat\lambda_i^*\hat\omega_i,
\end{align*}

\noindent where

\[
\hat\lambda_i^* = (1 - \hat\rho^*)\hat\lambda_i + \hat\rho^*.
\]

Thus, the vector of eigenvalues of $\hat\Omega^*$ is simply the vector of eigenvalues of $\hat\Omega$ shrunk toward the vector of ones with the exact same $\hat\rho^*$, and the orthonormal matrix of eigenvectors remains the same. This method of eigenvalue shrinkage possesses two important and intuitive properties (as opposed to the second-order bias correction described earlier), namely that $\hat\lambda^*_1 \geq \dots \geq \hat\lambda^*_p > 0$ and $\sum_{i=1}^p\hat\lambda^*_i = p.$ In addition, it is easy to see the reduction in variance from the shrinkage is

\[\Var(\hat\lambda^*) = \Var\big((1 - \hat\rho^*)\hat\lambda_i + \hat\rho^*\big) = (1 - \hat\rho^*)^2\Var(\hat\lambda).\]

In some situations $\hat\lambda^*$ is less biased than $\hat\lambda$, and in other situations the shrinkage increases bias. We explore this bias, along with other components of this paper, in the simulations below. 

\subsection{Second-Order Bias Correction}

In the following section we present the technical details for the second-order bias correction method, which is a generalization of the second-order approach presented in \cite{anderson1965asymptotic}.

\subsubsection{A Perturbation Theory Result}

We begin the development of our general bias correction term with a well known result from matrix perturbation theory. Derivations of this result can be easily found (see \cite{fukunaga2013introduction, sakai2000quadratic, saleem2015perturbation}) and so a proof is omitted here.

\begin{lem}\label{Lemma: SecondOrderPerturbation}
    Let $M_0$ be a $p\times p$ symmetric positive definite matrix with distinct eigenvalues $\lambda_1 > \dots > \lambda_p$, and corresponding unit eigenvectors $\omega_1, \dots, \omega_p$. Let $\hat{M} = M_0 + \delta V$ where $V$ is some symmetric random matrix with $\E[V] = 0$ and $\delta > 0$, and let $\hat\lambda_i$ be the $i$th eigenvalue of $\hat{M}$, with corresponding unit eigenvector $\hat\omega_i$. Then,
    
    \[
    \E[\hat\lambda_i] = \lambda_i + \sum_{i \neq j} \frac{\E[(\omega_i^T \hat{M} \omega_j)^2]}{\lambda_i - \lambda_j} + \mathcal{O}(\delta^3).
    \]
\end{lem}

The difficulty with using this equation to develop a bias correction comes from calculating the expectation in the numerator of the sum. However, with a simple transformation involving Lemma \ref{Lemma: KroneckerVec}, we can quickly apply this equation to reduce estimation bias. 

\subsubsection{A Plug-in Bias Adjustment}

We now provide a simple result which expresses the difficult expectation given in Lemma \ref{Lemma: SecondOrderPerturbation} in terms of the variance of the perturbed matrix.

\begin{lem}\label{Lemma: ExpectQuadSq}
    Let $\hat{M}$ be a random matrix with $E[\hat{M}] = M$ and $\Var[\vecT(\hat{M})] = \Sigma_{\hat M}$. Also, let $\omega_i$ be the $i$th eigenvector of $M$. Then for $i\neq j$,
    
    \[
        E[(\omega_i^T \hat{M} \omega_j)^2] = (\omega_j \otimes \omega_i)^T \Sigma_{\hat M}(\omega_j \otimes \omega_i).
    \]
\end{lem}

\noindent \textbf{Proof:}

\noindent Observe,

\begin{align*}
    \E[(\omega_i^T\hat{M}\omega_j)^2] &= \Var(\omega_i^T\hat{M}\omega_j) + (\E[\omega_i^T\hat{M}\omega_j])^2 \\
    &= \Var\big[(\omega_j^T\otimes\omega_i^T)\vecT(\hat{M})\big] + 0 \\
    &= (\omega_j \otimes \omega_i)^T \Var\big[\vecT(\hat{M})\big](\omega_j \otimes \omega_i) \\
    &= (\omega_j \otimes \omega_i)^T \Sigma_{\hat M} (\omega_j \otimes \omega_i).\blacksquare
\end{align*}

Lemma \ref{Lemma: ExpectQuadSq} allows us to use the perturbation expression from Lemma \ref{Lemma: SecondOrderPerturbation} as a plug-in bias correction for any transformation of the covariance matrix for which we can estimate the variance. This idea is formalized in the following Theorem, which was presented above in Section 3.2 without proof.

\begin{thm}\label{Theorem: BiasCorrection}
    Let $X$ be a $p \times 1$ random vector with finite moments up to the fourth order, and invertible covariance matrix $\Sigma$. Then, defining the unit eigenvectors of $\Omega$ as $\omega_1, \dots, \omega_p$ we have
    
    \[
        \E[\hat\lambda_i] = \lambda_i + \sum_{j \neq i}\frac{(\omega_j \otimes \omega_i)^T\Sigma_{\hat{\Omega}}(\omega_j \otimes \omega_i)}{\lambda_i - \lambda_j} + \mathcal{O}(n^{-\frac{3}{2}}).
    \]
\end{thm}

\noindent \textbf{Proof:}

\noindent Observe that from Lemma \ref{Lemma: SecondOrderPerturbation}, with $\delta = n^{-1/2}$, we must have that 

\[
\E[\hat\lambda_i] = \lambda_i + \sum_{i \neq j}\frac{\E\big[(\omega_j^T\hat{\Omega}\omega_i)^2\big]}{\lambda_i - \lambda_j} + \mathcal{O}(n^{-\frac32}),
\]

\noindent which then by Lemma \ref{Lemma: ExpectQuadSq} gives 

\[
    \E[\hat\lambda_i] = \lambda_i + \sum_{j \neq i}\frac{(\omega_j \otimes \omega_i)^T\Sigma_{\hat{\Omega}}(\omega_j \otimes \omega_i)}{\lambda_i - \lambda_j} + \mathcal{O}(n^{-\frac{3}{2}}).\blacksquare
\]

\newpage 

\section{}

In this paper we have suggested multiple techniques for approaching the estimation of the eigenvalues of the normalized inverse covariance matrix $\Omega$, along with a Stein-type shrinkage estimator for $\Omega$ itself. There are practical complications that come along with both of these problems that are most clearly seen through numerical simulations. Thus, in the following section we present simulations pertaining first to the estimation of the shrinkage parameter $\rho^*$, and then to the estimation of the eigenvalues of $\Omega$.

\subsection{The Shrinkage Parameter}

Our first simulation goal is to examine the bias induced by estimating $\rho^*$ using $\hat\rho^*.$ In order to accomplish this we generated data sets of dimension $p = 20$ and sizes $n = 30, 100, 250, 500$ under 4 different generative models. For model $i$ the data was generated so that $X \sim N(0, \Sigma^{(i)})$, where we chose $\Sigma^{(i)}$ so that the true largest eigenvalue of $\Omega^{(i)}$ was $\lambda_1^{(i)} = 2, 2.4, 5.8, 10.4.$ In each situation we generated $B = 300$ data sets and calculated $\hat\rho^*$ from each. 

\noindent Recall that Theorem \ref{Theorem: rhoStarConsistent} gives

\[\hat\rho^* = \frac{\widehat{\tr(\Sigma_{\hat\Omega})}}{\widehat{\tr(\Sigma_{\hat\Omega})} + \sum_{i=1}^p\hat\lambda_i^2 - p}.\]

\begin{figure}[t]
    \centering
    \includegraphics[width = .9\textwidth]{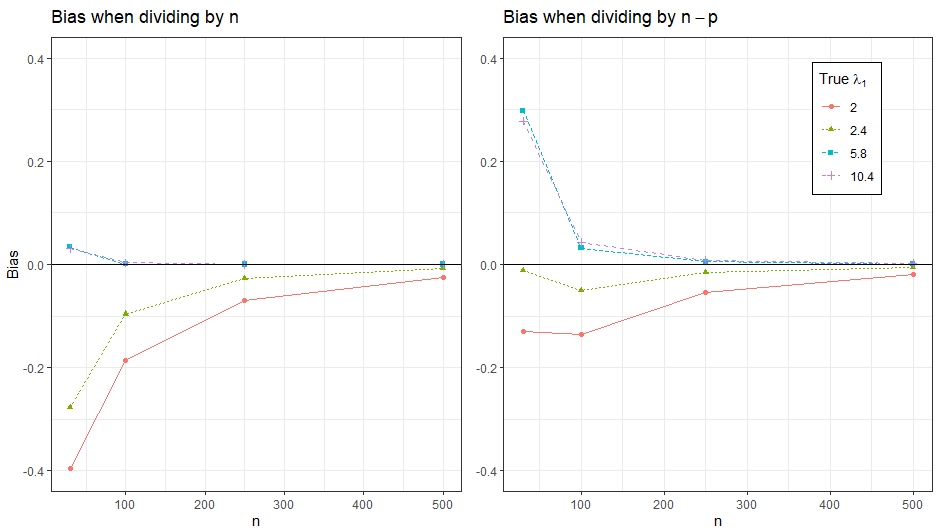}
    \caption{Bias in $\hat\rho^*$ when estimating the shrinkage parameter $\rho^*$. The left panel shows the bias when estimating $\hat\rho^*$ using a plug-in estimate of $\Sigma_{\hat\Omega}$ that divides by $n$, and the right shows the bias when using the same plug-in estimate but instead dividing by $n-p$. The different shapes and colors represent this bias calculated under different generative models, where the true $\lambda_1$ is given as in the legend. }
    \label{Figure: rhoStarBias}
\end{figure}

As mentioned above, it is reasonable to use $\widehat{\tr(\Sigma_{\hat\Omega})} = \tr(\hat{U}^T\hat{V}^*\hat{U})/n$ or $\widehat{\tr(\Sigma_{\hat\Omega})} = \tr(\hat{U}^T\hat{V}^*\hat{U})/(n-p)$ as in both situations we have a consistent estimator of $\tr(\Sigma_{\hat\Omega})$ and we have assumed $n > p$. However, as can be seen in Figure \ref{Figure: rhoStarBias} the choice of denominator has a significant impact on the bias in the estimation of $\hat\rho^*.$

The left panel of Figure \ref{Figure: rhoStarBias} shows the bias incurred when calculating $\hat\rho^*$ using $\widehat{\tr(\Sigma_{\hat\Omega})} = \tr(\hat{U}^T\hat{V}^*\hat{U})/n$. As can be clearly seen, in situations when $\lambda_1$ is larger (5.8 or 10.4 in our case), the small-sample bias is very small, and tends to slightly over-estimate $\rho^*$. However, in situations where $\lambda_1$ is smaller, then this particular estimate of $\rho^*$ severely under-estimates the true value in small samples. 

This can be constrasted with the right panel of Figure \ref{Figure: rhoStarBias}, which shows the bias when using $\widehat{\tr(\Sigma_{\hat\Omega})} = \tr(\hat{U}^T\hat{V}^*\hat{U})/(n-p)$. In this case we see that when we have larger values of $\lambda_1$, $\rho^*$ tends to be somewhat over-estimated in small samples, while with smaller values of $\lambda_1$ we see that $\rho^*$ is still under-estimated in small samples, but by a lesser margin. 

These simulations lead us to recommend the use of $n-p$ in the denominator of $\widehat{\tr(\Sigma_{\hat\Omega})}$, as in most situations it is more helpful to over-shrink rather than over-fit. However, if an investigator has reason to believe that in their situation $\lambda_1$ is relatively large or desires to avoid over-shrinkage, then using $n$ in the denominator is also justified. 

Either way, it is clearly seen that as the sample size increases the bias disappears, reinforcing the claim from Theorem \ref{Theorem: rhoStarConsistent} that $\hat\rho^*$ is consistent.

\subsection{The Eigenvalues of $\Omega$}

We have presented several different estimators for the eigenvalues of $\Omega$. These are the sample, the second-order corrected, shrinkage and second-order shrinkage eigenvalues, denoted $\hat\lambda, \hat\lambda^c, \hat\lambda^*,$ and $\hat\lambda^{c*}$ respectively. In the following subsection we present simulations exploring the performance of these 4 different estimators at various sample sizes, and under various true models. We primarily focus on the estimation of the largest eigenvalue in these simulations, although we also present some results regarding the smallest eigenvalue.

As in the shrinkage simulations, we generated data sets of dimension $p = 20$ and sizes $n = 30, 100, 250, 500, 1000$ under 5 different generative models. For model $i$ the data was generated so that $X \sim N(0, \Sigma^{(i)})$, where we chose $\Sigma^{(i)}$ so that the largest eigenvalue of $\Omega^{(i)}$ was $\lambda_1^{(i)} = 1, 2, 2.4, 5.8, 10.4.$ In each situation we generated $B = 300$ data sets and calculated $\hat\lambda, \hat\lambda^c, \hat\lambda^*,$ and $\hat\lambda^{c*}$ from each. 

\begin{figure}[ht]
    \centering
    \begin{minipage}{0.45\textwidth}
        \centering
        \includegraphics[height=6cm]{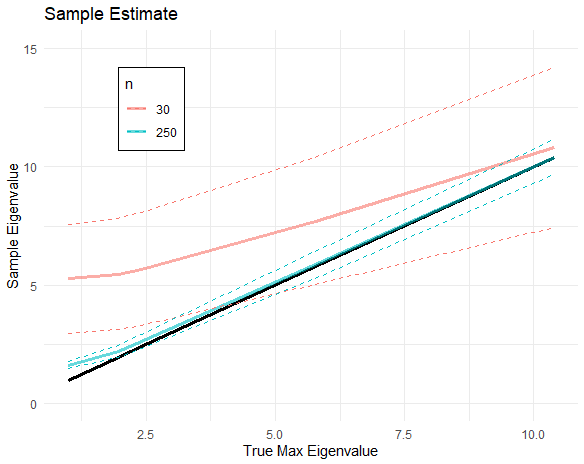} 
    \end{minipage}\hspace{.2cm}
    \begin{minipage}{0.45\textwidth}
        \centering
        \includegraphics[height=6cm]{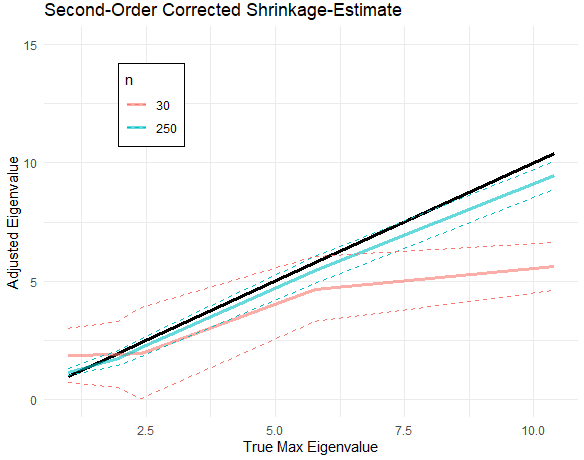} 
    \end{minipage}
    \caption{The difference in the bias of estimators $\hat\lambda_1$ and $\hat\lambda^{c*}_1$ (sample and second-order corrected shrinkage respectively) as a function of the true value of $\lambda_1.$ The black line shows the true $\lambda_1$, while the red and blue lines show the average estimated value for $n = 30$ and $n = 250$ respectively. The dashed lines give the 95\% point-wise confidence bands. The left panel shows the results for $\hat\lambda$, while the right shows the results for $\hat\lambda^{c*}.$}
    \label{Figure: sampEigenStackedBias}
\end{figure}

One of the interesting aspects of estimating the largest eigenvalue of $\Omega$ is that the sampling bias generally \textit{decreases} as the true maximum eigenvalue increases (holding both $n$ and $p$ constant). This is demonstrated in the left panel of Figure \ref{Figure: sampEigenStackedBias}. As can be seen, the bias increases with smaller values of $\lambda_1$ to the point that the true value lies outside the confidence interval, even at a higher sample size. 


This problem remains for smaller values of $\lambda_1$ even when using the shrinkage estimator $\hat\lambda_1^*$. However, as shown in the right panel of Figure \ref{Figure: sampEigenStackedBias}, using the second-order bias correction in combination with the shrinkage can be a successful technique for accurate inference on $\lambda_1$ when the true value is smaller. Of course, the shrinkage bias prevents accurate inference on $\lambda_1$ when the true value is larger, providing a trade off for investigators to consider.

\begin{figure}[ht]
    \centering
    \includegraphics[width = .8\textwidth]{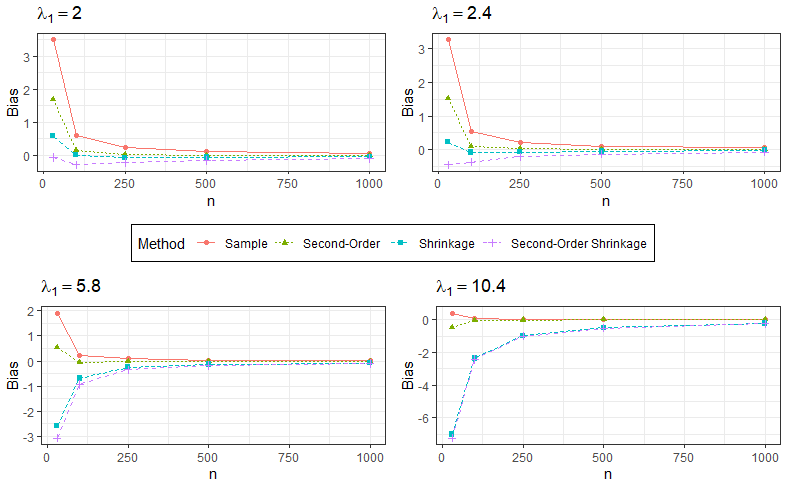}
    \caption{This figure presents the average Monte Carlo bias across simulations while estimating $\lambda_1$. Each panel represents a different generative model with the true value listed above, the x-axis for each gives the sample size, the y-axis gives the bias, and the different lines/shapes represent different estimation methods.}
    \label{Figure: maxLambdaBias}
\end{figure}

Finally, we present bias results from our simulations which demonstrate the varied ways in which the different bias correction methods react under different generative models. Figure \ref{Figure: maxLambdaBias} shows the average Monte Carlo bias while estimating $\lambda_1$ for each of the 4 methods presented in this paper, at multiple different sample sizes, and for the different generative models. As can be seen, for smaller true values of $\lambda_1$ the shrinkage and second-order shrinkage methods work best, while for larger true values it is best to avoid shrinkage entirely. 

As discussed above, we are interested in the use of $\lambda_1$ in the hypothesis test with null and alternative hypotheses

\begin{align*}
    &H_0: \lambda_1 \leq 2 &H_A: \lambda_1 > 2.
\end{align*}

The results shown in Figure \ref{Figure: maxLambdaBias} clearly indicate that under the null hypothesis ($\lambda_1 \leq 2$), the best estimation method is the second-order shrinkage approach as this is the only approach with no positive bias, which in the case of our hypothesis test, would artificially inflate the Type I error.

\begin{figure}[ht]
    \centering
    \includegraphics[width = .8\textwidth]{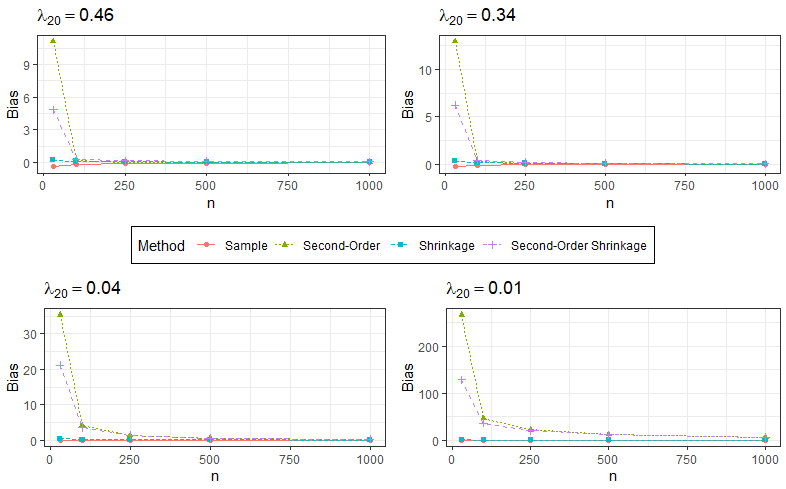}
    \caption{This figure presents the average Monte Carlo bias across simulations while estimating $\lambda_{20}$. Each panel represents a different generative model with the true value listed above, the x-axis for each gives the sample size, the y-axis gives the bias, and the different lines/shapes represent different estimation methods.}
    \label{Figure: minLambdaBias}
\end{figure}

While estimating $\lambda_{20}$ (the smallest eigenvalue), Figure \ref{Figure: minLambdaBias} shows a significant amount of instability in the second-order correction method, as expected. Thus, we do not recommend using this technique unless the user is able to assume that the target eigenvalue is adequately separated from the others in the true model and in the sample. This assumption is generally most likely to be fulfilled in the largest eigenvalue, but can also be frequently fulfilled for the second largest or other higher rank eigenvalues. Typically however, the second-order bias correction method should not be used for the smallest or other lower rank eigenvalues as those tend to cluster together, at least in the sample.

\clearpage
\bibliography{bibliography}

\end{document}